\documentclass[10pt,preprint]{article}
\PassOptionsToPackage{table,xcdraw}{xcolor}  % 
\let\origaddcontentsline\addcontentsline

\usepackage{rlc}

\let\addcontentsline\origaddcontentsline

\usepackage{hyperref}
\hypersetup{
    colorlinks, linkcolor={dark-blue},
    citecolor={dark-blue}, urlcolor={dark-blue}
}

\usepackage{etoolbox}
\pretocmd{\section}{\phantomsection}{}{}
\pretocmd{\subsection}{\phantomsection}{}{}
\usepackage{tikz}
\usepackage{tcolorbox}
\usepackage{xcolor}
\usetikzlibrary{shapes,arrows,positioning,decorations.pathmorphing}

\usepackage{authblk} % For affiliation formatting
\usepackage{amssymb}            % Defines common symbols like \mathbb R
\usepackage{mathtools}          % Extends amsmath, providing common math tools
\usepackage{mathrsfs}           % Enables \mathscr, which can work in cases that \mathcal does not
\usepackage{amsthm}
\newtheorem{definition}{Definition}
\mathtoolsset{showonlyrefs}     % Only number equations that are referenced (optional)
\usepackage{graphicx}           % For including images
\usepackage{subcaption}         % Allows for the use of subfigures and subcaptions
\usepackage[space]{grffile}     % For spaces in image names
\usepackage{url}                % For displaying urls
\usepackage{booktabs}       % professional-quality tables
\usepackage{amsfonts}       % blackboard math symbols
\usepackage{nicefrac}       % compact symbols for 1/2, etc.
\usepackage{enumitem}  % 
\usepackage{graphicx}  % for resizing
\usepackage{longtable}  % to allow for tables that span multiple pages
\usepackage{array}  % for better table formatting
\usepackage{tabularx}
\usepackage{float}
\usepackage{placeins}
\usepackage{tikz}
\usepackage{tikz-qtree}
\usetikzlibrary{trees,positioning,shapes,arrows}
\usepackage{natbib}
\usepackage{doi}
\usepackage{svg}
\usepackage{tikz}
\usepackage[table]{xcolor}
\usetikzlibrary{trees, shapes, arrows, positioning, calc}
\usepackage{array, makecell, booktabs}
\usepackage{ragged2e}
\newcolumntype{P}[1]{>{\RaggedRight\arraybackslash}p{#1}}
\definecolor{greyboxbg}{RGB}{221,221,221}
\definecolor{blueboxbg}{RGB}{240,240,240}
\definecolor{orangeboxbg}{RGB}{200,255,200}
\definecolor{greenboxbg}{RGB}{142,207,201}
\usepackage{multirow} 
\usepackage{tikz}
\usetikzlibrary{shadings,arrows.meta}

\usepackage{marvosym}

\usepackage{tocloft}
\usepackage[export]{adjustbox}
\usepackage{xcolor}

\usepackage{tikz}
\usetikzlibrary{positioning,calc}
\usepackage{xcolor}
\usepackage{graphicx}

\definecolor{PRSIblue}{RGB}{219,237,250}
\definecolor{PRSIblueLine}{RGB}{67,126,170}

\definecolor{HTVgreen}{RGB}{224,244,232}
\definecolor{HTVgreenLine}{RGB}{72,153,112}

\definecolor{LeafPurple}{RGB}{239,228,249}
\definecolor{LeafPurpleLine}{RGB}{132,92,166}

\definecolor{PaperYellow}{RGB}{255,249,218}
\definecolor{PaperYellowLine}{RGB}{211,176,63}

\definecolor{TaxonomyLine}{RGB}{65,65,65}

\usepackage[table]{xcolor}
\usepackage{booktabs}
\usepackage{tabularx}
\usepackage{array}
\usepackage{makecell}

\definecolor{PRP0}{HTML}{EDF4F8}
\definecolor{PRP1}{HTML}{E4EFF6}
\definecolor{PRP2}{HTML}{D9EAF3}

\definecolor{PRR}{HTML}{E8F3EE}
\definecolor{PRCell}{HTML}{FAFBFC}

\definecolor{PRphysical}{HTML}{247A92}
\definecolor{PRdigital}{HTML}{5F6872}
\definecolor{PRuncertain}{HTML}{B7791F}
\definecolor{PRmuted}{HTML}{8A949E}

\usepackage[edges]{forest}
\usepackage{tikz}
\usetikzlibrary{arrows.meta}

\tcbset{
  examplebox/.style={
    colback=gray!5,
    colframe=gray!40,
    fonttitle=\bfseries,
    title=Example,
    coltitle=black,
    boxrule=0.4pt,
    arc=2mm,
    top=1mm,
    bottom=1mm,
    left=1mm,
    right=1mm,
  }
}
\newtcolorbox{example}[1][]{examplebox,#1}

\tcbset{Conditionbox/.style={
  colback=blue!3!white,
  colframe=blue!60!black,
  coltitle=black,
  title= Core Conditions for the Agentic Web.
}}

\tcbset{definitionbox/.style={
  colback=blue!5!white,
  colframe=blue!75!black,
  fonttitle=\bfseries,
  coltitle=white,
  title=Definition: Agentic Web
}}

\tcbset{
  myboxstyle/.style={
    fonttitle=\bfseries,
    fontupper=\normalsize,
    boxrule=0.6pt,
    arc=1.5mm,
    top=2mm,
    bottom=2mm,
    left=2mm,
    right=2mm,
  }
}

\tcbset{keyterminology/.style={
  colback=blue!3!white,
  colframe=blue!60!black,
  coltitle=black,
  title=Conceptual Layers,
  colbacktitle=blue!20!white,
  myboxstyle
}}

\tcbset{conceptrelation/.style={
  colback=blue!3!white,
  colframe=blue!60!black,
  coltitle=black,
  title=Core Conditions for the Agentic Web,
  colbacktitle=blue!20!white,
  myboxstyle
}}

\tcbset{frameworkoverview/.style={
  colback=blue!3!white,
  colframe=blue!60!black,
  coltitle=black,
  title=Framework Overview,
  colbacktitle=blue!25!white,
  myboxstyle
}}

\tcbset{example/.style={
  colback=blue!3!white,
  colframe=blue!60!black,
  coltitle=black,
  title=Example (Transactional),
  colbacktitle=blue!20!white,
  myboxstyle
}}

\tcbset{example2/.style={
  colback=blue!3!white,
  colframe=blue!60!black,
  coltitle=black,
  title=Example (Informational),
  colbacktitle=blue!20!white,
  myboxstyle
}}

\tcbset{layerinteraction/.style={
  colback=blue!3!white,
  colframe=blue!60!black,
  coltitle=black,
  title=Layer Interaction,
  colbacktitle=blue!22!white,
  myboxstyle
}}

\usepackage[edges]{forest}
\usepackage{adjustbox}
\usepackage{xcolor}

\definecolor{verroot}{HTML}{28346F}

\definecolor{verartifact}{HTML}{8C8C8C}
\definecolor{verartifactlight}{HTML}{F0F0F0}

\definecolor{verobs}{HTML}{E7A72E}
\definecolor{verobslight}{HTML}{FFF1CF}

\definecolor{verpred}{HTML}{3FA7D6}
\definecolor{verpredlight}{HTML}{E2F2FA}

\definecolor{verrelease}{HTML}{36A98F}
\definecolor{verreleaselight}{HTML}{E0F4EF}

\definecolor{vaRoot}{HTML}{3C3B6E}
\definecolor{vaAdapt}{HTML}{D99152}
\definecolor{vaAdaptLight}{HTML}{FBE9D8}
\definecolor{vaEval}{HTML}{5798C5}
\definecolor{vaEvalLight}{HTML}{E3F0F8}
\definecolor{vaAudit}{HTML}{55A184}
\definecolor{vaAuditLight}{HTML}{E3F3EC}
\definecolor{vaCriterion}{HTML}{9A78B5}
\definecolor{vaCriterionLight}{HTML}{EEE7F4}

\newtoggle{final}

\togglefalse{final} % comment this line for final version

\usepackage{enumitem}

\title{Physical Recursive Self-Improvement}

\title{Who Improves the Robot? Toward Physical Recursive Self-Improvement}

\author{%
\parbox{\textwidth}{%
\centering

{\linespread{1.2}\selectfont
\name{Hai Zhou}$^{1,*,\dagger}$,
\name{Fan Yang}$^{1,*}$,
\name{Zeyi Chen}$^{1,*}$,
\name{Lejun Zhang}$^{1,*}$,
\name{Chenyang Fu}$^{1,*,\dagger}$,
\name{Jianping Li}$^{2}$,
\name{Jiayi Zhu}$^{1}$,
\name{Xianyang Liu}$^{1}$,
\name{Yingxuan Yang}$^{1}$,
\name{Lili Meng}$^{2}$,
\name{Lu Wang}$^{3}$,
\name{Xinyang Han}$^{4}$,
\name{Jiankai Sun}$^{5}$,
\name{Weirui Ye}$^{6}$,
\name{Guanchu Wang}$^{7}$,
\name{Wanhua Li}$^{8}$,
\name{Ming Zhou}$^{7}$,
\name{Boyi Li}$^{9}$,
\name{Ming Jin}$^{10}$,
\name{Kun Shao}$^{11}$,
\name{Muhao Chen}$^{12}$,
\name{Guang Chen}$^{13}$,
\name{Lei Ma}$^{14}$,
\name{Joey Tianyi Zhou}$^{15}$,
\name{Costas Spanos}$^{4}$,
\name{Xia Hu}$^{7}$,
\name{Mac Schwager}$^{5}$,
\name{Weinan Zhang}$^{1,}$\textsuperscript{\Letter},
\name{Qian Feng}$^{16,}$\textsuperscript{\Letter},
\name{Shangding Gu}$^{1,4,}$\textsuperscript{\Letter}

\par}

\vspace{8pt}

$^1$Shanghai Jiao Tong University\par\vspace{2pt}
$^2$Google  \hspace{20pt}
$^3$Microsoft \par\vspace{2pt}
$^4$University of California, Berkeley\par\vspace{2pt}
$^5$Stanford University\par\vspace{2pt}
$^6$Massachusetts Institute of Technology\par\vspace{2pt}
$^7$Shanghai Artificial Intelligence Laboratory\par\vspace{2pt}
$^8$Nanyang Technological University \par\vspace{2pt}
$^{9}$Nvidia  \hspace{15pt}
$^{10}$Virginia Tech \hspace{15pt}
$^{11}$Xiaomi  \par\vspace{2pt}
$^{12}$University of California, Davis\par\vspace{2pt}
$^{13}$Tongji University \par\vspace{2pt}
$^{14}$University of Tokyo \par\vspace{2pt}
$^{15}$Singapore Management University \par\vspace{2pt}
$^{16}$Technical University of Munich\par

}%
}

\begin{document}

\begingroup
\renewcommand{\thefootnote}{\ensuremath{*}}
\footnotetext{Equal contribution.}
\endgroup

\begingroup
\renewcommand{\thefootnote}{\ensuremath{\dagger}}
\footnotetext{Intern at Shanghai Jiao Tong University.}
\endgroup

\begingroup
\renewcommand{\thefootnote}{\Letter}
\footnotetext{Correspondence to: \textit{shangding.gu@sjtu.edu.cn}; \textit{qian.feng@tum.de}; \textit{wnzhang@sjtu.edu.cn}.}

\endgroup

\maketitle

\vspace{-0.6cm}

\begin{abstract}

Robots are increasingly autonomous in task execution, yet their improvement remains largely human-driven. This study investigates Physical Recursive Self-Improvement (\textit{\textbf{Physical RSI}}), in which physical experience guides lasting updates to both a robot's capabilities and its improvement procedure, making subsequent improvement more effective across tasks. Physical RSI faces challenges from noisy observations, continuous dynamics, costly trials, and irreversible consequences. We present an L0--L5 framework that differentiates capability-level from meta-level updates and within-task from across-task gains, reserving L5 for independently verified recursive gain. We study existing methods around harness adaptation and model training, alongside safety and control mechanisms that govern execution and successive updates. Our analysis reveals that, despite substantial progress in self-improvement, the surveyed literature does not yet demonstrate Physical RSI as defined by our framework. We identify research priorities in reliable feedback, causal attribution, safety preservation, and benchmarking, outlining a path toward robots that autonomously improve how they learn while ensuring safety. The GitHub repository for Physical RSI is available at \url{https://github.com/SAIL-Research-Lab/Physical-RSI}.
\end{abstract}


\newpage
\tableofcontents

\newpage

\newcommand{\Train}{\mathsf{Train}}
\newcommand{\Ver}{\mathsf{Ver}}
\newcommand{\Adapt}{\mathsf{Adapt}}

\section{Introduction}
\label{sec:intro}
 
Advances in robotics have enabled machines to carry out an expanding range of tasks on their own, from sorting parcels in logistics centers to performing routine chores at home \citep{zeng2018robotic,fu2024mobile}. Yet behind this execution autonomy lies a reliance on human support. In data preparation, robots do not have data as abundant as the internet data available to language models. Their data is generated through thousands of physical interactions, and humans are needed to collect and review it \citep{openx2024,khazatsky2024droid}. In deployment, when a robot meets an unfamiliar failure, recovery largely depends on humans to diagnose the problem or demonstrate a fix \citep{liu2023sirius}. In training and optimization, most systems rely on training pipelines designed by humans, and the final choice of which update to adopt remains a human decision \citep{zhu2020ingredients}. Consequently, even highly capable robots lack \emph{improvement autonomy}, meaning the ability to improve themselves without humans driving the loop.

Recent work has begun to narrow this gap by letting systems improve themselves. One line of work lets language model agents rewrite code or scaffolding, and keep the changes that score better under automated evaluation \citep{zelikman2023stop,yin2025godel,zhang2025dgm,novikov2025alphaevolve}. Another line improves the model itself, training it on data and feedback that the system or another language model generates \citep{yuan2024selfrewarding,zweiger2025seal,ghasemipour2025selfimproving,ma2024eureka,wang2024robogen}. Most of this work achieves self-improvement, in that the system raises its own task performance by changing what it uses when it performs a task. The procedure that produces and selects these changes usually stays fixed, so what the system can gain is limited by how effective that procedure is. Recursive self-improvement goes further. It updates the improvement procedure itself. Recursive gain is established when the revised procedure produces better subsequent improvements than its predecessor under matched conditions. Few works however update the procedure, and fewer show that doing so yields recursive gain.

Moreover, recursive self-improvement is harder for robots in the real world than in simulated or digital environments. The physical world cannot be monitored as exhaustively as digital environments. A robot observes only partial and noisy traces of what actually happened, which makes failures hard to diagnose and improvements hard to verify \citep{dulacarnold2019challenges}. Outcomes are also less deterministic than in digital environments, and the robot must act through continuous dynamics under real-time constraints \citep{dulacarnold2019challenges}. Evaluating any change therefore requires repeated trials under noisy conditions, which is time-consuming on real robots \citep{zhou2025autoeval}. Finally, resetting a physical system is expensive, hardware can fail, and many interactions risk irreversible outcomes, including harm to people who share the workspace, so the cheap trial and error on which self-improvement in digital environments relies becomes costly and risky \citep{eysenbach2018leave,zhu2020ingredients}. These constraints make it particularly hard on real robots to update and to verify the changes. Physical robots therefore call for a recursive self-improvement framework of their own, rather than one carried over from digital settings.

We build this framework around \emph{Physical Recursive Self-Improvement} (Physical RSI). We say that a robot performs Physical RSI when it uses evidence from its physical interaction to persistently update both its task-execution system and its improvement procedure, such that its subsequent improvement becomes more effective over successive rounds. Here the \emph{task-execution system} is the models and harness that the robot uses to perform tasks, and the \emph{improvement procedure} is the part that can update the task-execution system and itself. An independent evaluator assesses the gains of the updated system. Our long-term vision is for robots to operate and evolve autonomously in real-world physical environments, addressing diverse and unforeseen challenges without human intervention.

\begin{figure*}[t]
    \centering
    \includegraphics[width=\textwidth]{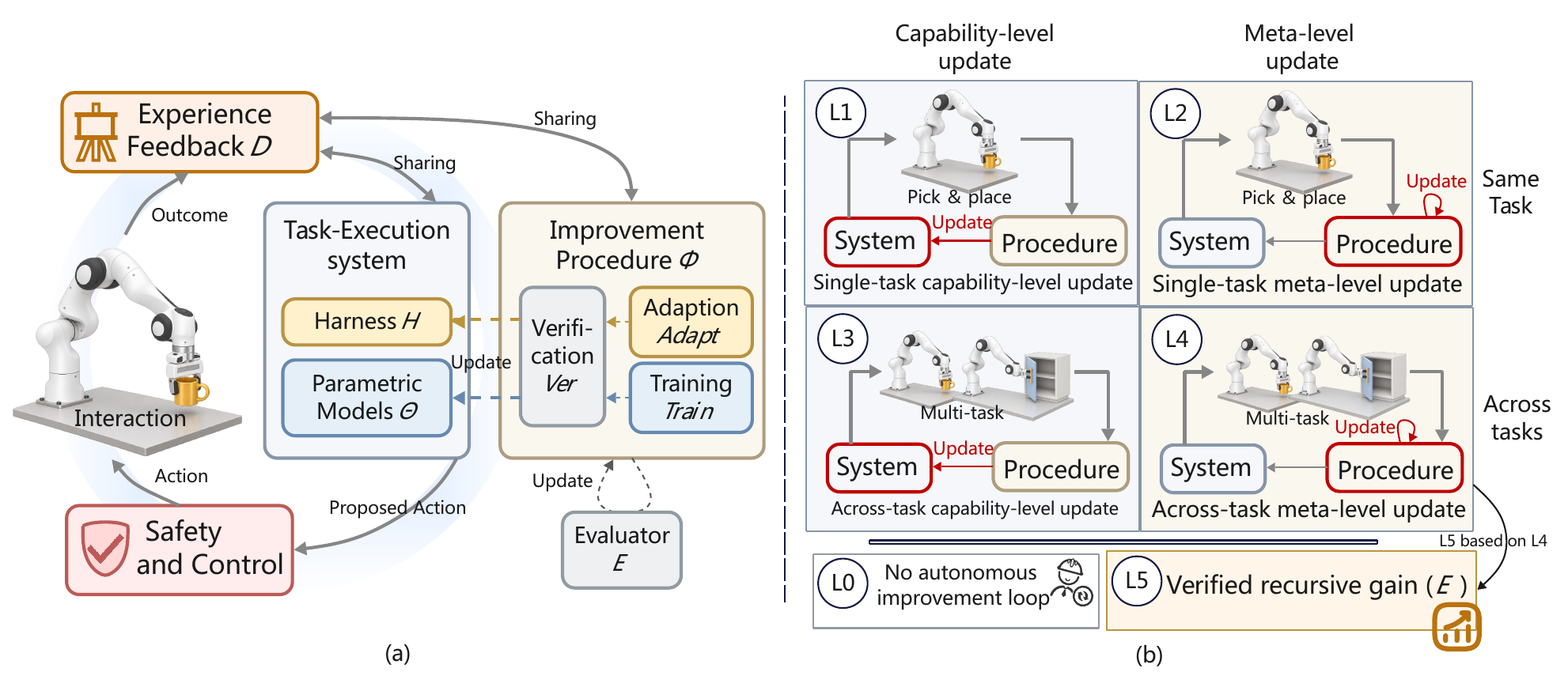}
    \caption{Overview of Physical Recursive Self-Improvement. (a) The task-execution system, consisting of the harness $H$ and the parametric models $\Theta$, proposes actions that pass through safety and control before acting on the physical world. The outcomes are stored as experience and feedback $D$, which is shared by the task-execution system and the improvement procedure $\Phi$. Using $D$, $\Phi$ updates $H$ through $\Adapt$ and $\Theta$ through $\Train$, while $\Ver$ decides which changes are kept. $\Phi$ can also update its own stages. An independent evaluator $E$ evaluates the task-execution system and thereby judges the updates made by $\Phi$. (b) L1--L4 distinguish capability-level updates to the task-execution system from meta-level updates to the improvement procedure, within a single task or across tasks. L0 denotes systems that make no retained change on their own. L5 builds on L4 and requires independent, controlled evidence that subsequent improvement becomes more effective over successive rounds.}
    \label{fig:overview}
\end{figure*}

Based on this definition, we introduce six levels, L0 to L5, to describe how much of it a system meets, as shown in Figure~\ref{fig:overview}. L0 covers systems that make no retained change on their own, either because they adjust only within a run or because humans still drive the loop. L1 to L4 vary along two axes. Capability-level updates change the task-execution system, whereas meta-level updates change the improvement procedure. The second axis is scope. A gain on a single task may be specific to that task, whereas a gain across multiple tasks shows that the update generalizes. L1 makes capability-level updates on a single task with the improvement procedure fixed, and L2 adds meta-level updates on the same task. L3 and L4 extend L1 and L2, respectively, to multiple tasks. L5 builds on L4 and requires evidence, from an independent evaluator under the same controlled conditions, that each update yields a larger gain than the previous one. The formal definition and the detailed criteria are given in Section~\ref{sec:foundations}.

Existing surveys cover complementary parts of this framework. Surveys centered on components and experience examine how agents change their models, prompts, memory, tools, and workflows \citep{gao2025survey,fang2025comprehensive,ren2026selfimprovements,jiang2026selfimproving}. RSI-oriented accounts ask whether the mechanisms that generate, evaluate, and select updates can themselves change \citep{liu2026path,duan2026last,chen2026recursive,wang2026diving}, and address safety through governance, independent evaluation, and the preservation of safeguards across updates \citep{liu2026path,wang2026diving,gong2026evolutionary}. For humanoids, \citet{nguyen2026humanoids} formulate admissible evolution through layered safety mechanisms. Such safeguards carry more weight in the physical world, where a failed trial cannot simply be rerun or rolled back \citep{duan2026last,nguyen2026humanoids}. We therefore make safety and control explicit in our system framework and dedicate a section to how they constrain physical interaction and autonomous improvement. And existing accounts characterize the scope of improvement in different ways. \citet{liu2026path} reserve a higher level for broad transfer, \citet{wang2026diving} treat transfer as a separate empirical question, and \citet{wu2026aiforai} separately record improvement targets and held-out transfer while cautioning against collapsing independent dimensions into a single autonomy ladder. For robots, generalization remains uneven across changes in objects, embodiment, and deployment conditions \citep{hou2026evaluation}. We therefore cross two criteria in L0--L5, namely, whether an update changes the task-execution system or the improvement procedure, and whether its demonstrated effects remain within one task or extend across tasks. Finally, the task-execution system and the improvement procedure play different roles, but both involve models, harnesses, and associated improvement procedures \citep{liu2026path,ren2026selfimprovements,wu2026aiforai}. We therefore organize methods into harness adaptation and model learning. This groups methods that operate on similar components while preserving the distinction between what the robot can do and how it improves.

This study makes three contributions. First, we propose Physical RSI as a concept and give it a graded framework from L0 to L5. The framework treats what is updated and how far an update reaches as orthogonal criteria to define L1--L4. Second, we investigate existing methods along two implementation paths, harness adaptation and model learning, each covering both the task-execution system and the improvement procedure, together with the safety and control mechanisms that bound both, and the verification that decides which updates are kept. Third, we review how existing systems are evaluated, set out a protocol for independently evaluating improvement procedures, and identify the evidence gaps that separate current work from higher levels.

The remainder of the study is organized as follows. Section~\ref{sec:foundations} formalizes the framework, including the definition, the system architecture, the L0--L5 levels, and the boundaries of evidence. Section~\ref{sec:harness} surveys harness adaptation, and Section~\ref{sec:training} surveys model learning. Section~\ref{sec:safety} addresses safety and control, and Section~\ref{sec:evaluation} examines how Physical RSI is evaluated. Section~\ref{sec:challenges} discusses open problems, and Section~\ref{sec:conclusion} concludes.

\section{Foundations of Physical Recursive Self-Improvement}
\label{sec:foundations}

This section establishes the foundations for analyzing Physical RSI. Section~\ref{sec:definition} defines the system components and the conditions for physical self-improvement and Physical RSI, and relates them to neighboring research paradigms. Section~\ref{sec:architecture} describes the execution, capability-level update, and meta-level update loops. Section~\ref{sec:levels} specifies task identity and the criteria for assigning levels L0--L5. This framework also provides a shared terminology for the analysis throughout the study.
 
\subsection{Definition and Scope of Physical RSI}
\label{sec:definition}
 
We describe an improving robot by its task-execution system and improvement procedure (Figure~\ref{fig:overview}(a)), together with two properties of its updates, physical grounding and inheritance.
 
\paragraph{Task-execution system} The task-execution system is the pipeline through which the robot turns a task into action. It has two parts. The \emph{harness} $H$ comprises the components outside the learned models, such as prompts, memory, skills and tools, task decomposition and recovery logic, and interface and controller code. The \emph{parametric models} are the learned models used during execution, such as a planner, a vision-language-action policy, or a value function or world model, and we denote their weights by $\Theta$. Capability can reside in either part, so we track both. Anything used only in the improvement procedure and not during task execution belongs to $\Phi$ rather than to the task-execution system (Section~\ref{sec:levels}).
 
\paragraph{Improvement procedure} The improvement procedure $\Phi$ is responsible for improving the whole system, including the task-execution system and the procedure itself. It has three stages. An adaptation stage $\Adapt$ updates $H$. A training stage $\Train$ selects experience and optimizes $\Theta$. A verification stage $\Ver$ judges each candidate change proposed by $\Adapt$ or $\Train$ and decides whether to keep it. $\Phi$ can also improve itself by changing these stages, with an update to $\Adapt$ or $\Train$ judged by $\Ver$ and an update to $\Ver$ judged by the independent evaluator $E$ introduced below. This matters because an improvement procedure is designed around particular tasks. When new tasks arrive, a fixed procedure usually has to be re-engineered by people, whereas one that can update itself adapts without them. We call changes to the task-execution system capability-level updates and changes to the improvement procedure meta-level updates, and Section~\ref{sec:architecture} describes how each is carried out.

\paragraph{Physical grounding} The improvement procedure draws its changes from experience, which it keeps in an experience and feedback store $D$. $D$ stores the records of task execution, together with the records of improvement. Memory in $H$ holds what $\Adapt$ has distilled from $D$ for use in $H$. Experience may come from the real world, from a physics simulator, from a learned world model, from humans, or from signals inside the system. Among these sources, only the real world shows how the robot's actions actually play out. A physics simulator reproduces contact and dynamics only approximately. A learned world model, humans, or internal signals can propose candidates or supply additional data, but a change drawn only from them is not physically grounded. We therefore call a change \emph{physically grounded} when it depends on what happened as the robot acted, either in the real world or in a physics simulator.

\paragraph{Inheritance} A change becomes the starting point of later behavior only if it is inherited. We call the state retained after $k$ rounds of execution and improvement the $k$-th generation $\mathcal{S}_k$,
\begin{equation}
\mathcal{S}_k=(\Theta_k,\ H_k,\ \Phi_k,\ D_k),\qquad \Phi_k=(\Adapt_k,\ \Train_k,\ \Ver_k),
\label{eq:state}
\end{equation}
in which $\Theta_k$ and $H_k$ form the task-execution system, $\Adapt_k$, $\Train_k$, and $\Ver_k$ form the improvement procedure, and $D_k$ is the experience and feedback store. Processes outside the robot, such as a training server or a language-model agent that edits code, belong to the system when a stage of $\Phi$ invokes them and they write into $\mathcal{S}_k$. In round $k+1$, generation $k$ executes tasks, its trajectories are added to $D_k$ to give $D_{k+1}$, and the improvement procedure $\Phi_k$, applied as the map defined by its three stages, produces the next generation,
\begin{equation}
(\Theta_{k+1},H_{k+1},\Phi_{k+1})=\Phi_k\left(\Theta_k,H_k,\Phi_k;\,D_{k+1}\right).
\label{eq:round}
\end{equation}
The initial generation $\mathcal{S}_0$ is the state at the start of the period over which improvement is claimed. A change is \emph{inherited} when $\Phi$ keeps it on evidence in $D$ and it persists beyond the episode in which it was made, so that later generations start from it unless a later update changes it. Since $\Phi_k$ is one of its own arguments in Eq.~\eqref{eq:round}, an inherited change may reach the improvement procedure, including $\Ver$ itself. A change to $\Ver$ cannot be judged by $\Ver$ alone, which is why the framework also needs a judge $E$ outside $\Phi$.
 
\paragraph{Verification and evaluation} $\Ver$ and the independent evaluator $E$ both judge changes, but they stand in different positions and serve different purposes. $\Ver$ is a stage of $\Phi$. It acts within every round, deciding which candidate changes proposed by $\Adapt$ or $\Train$ are kept, and because it belongs to $\Phi$, it can itself be changed. $E$ is a set of tasks with a scoring criterion $Q$ that $\Phi$ cannot read, train on, or modify, and it stands outside the system. It has two roles. It judges changes to $\Ver$, since a $\Ver$ that judged its own changes could learn to satisfy its own standard without becoming better. It also scores the task-execution system $(\Theta_k, H_k)$ once an update is complete, because it is this part that acts on the environment, while the improvement procedure is evaluated indirectly, by scoring the systems it produces. We write $J_Q(\Theta_k,H_k)$ for the score of generation $k$ on $E$ under $Q$. $E$ is extended from outside the system as it meets new tasks. Once $E$ is extended, earlier scores no longer measure the same thing, so earlier generations are either re-evaluated on the updated $E$ or compared on an anchor subset held fixed throughout the comparison. Section~\ref{sec:evaluation} describes how $E$ is built and used.
 
\paragraph{Definitions} Building on the task-execution system, the improvement procedure, physical grounding, inheritance, and the independent evaluator, we now define two notions of self-improvement. Definition~\ref{def:psi} asks whether the robot has become better through changes that it made and kept on its own, and Definition~\ref{def:prsi} asks, more strictly, whether the way it becomes better has also become better.

\begin{definition}[Physical self-improvement]
\label{def:psi}
A system performs physical self-improvement over rounds $1,\dots,K$ if the following conditions hold.
\begin{enumerate}
\item[(i)] \emph{Capability-level self-inheritance.} At least one change to $\Theta$ or $H$ is produced by stages of $\Phi$ and inherited. Changes to $\Phi$ may also be inherited but are not required. No human writes into $\Theta$, $H$, or $\Phi$.
\item[(ii)] \emph{Physical grounding.} The inherited changes on which the retained gain depends are physically grounded, whether in a physics simulator or in the real world.
\item[(iii)] \emph{Retained gain.} $J_Q(\Theta_K,H_K) > J_Q(\Theta_0,H_0)$.
\end{enumerate}
\end{definition}

\begin{definition}[Physical recursive self-improvement]
\label{def:prsi}
Physical self-improvement is recursive if the following conditions hold, where (i) and (iii) strengthen the corresponding conditions of Definition~\ref{def:psi}.
\begin{enumerate}
\item[(i)] \emph{Meta-level self-inheritance.} In addition to condition (i) of Definition~\ref{def:psi}, $\Phi$ is able to update its own stages, so that changes to $\Phi$ produced by $\Phi$ can be inherited and used by later rounds.
\item[(ii)] \emph{Physical grounding.} The inherited changes on which the gain depends are physically grounded, whether in a physics simulator or in the real world.
\item[(iii)] \emph{Recursive gain across tasks.} Starting from the same generation and experience, and under the same budget, the system produced by each updated $\Phi_k$ scores a higher $J_Q$ than the system produced by $\Phi_{k-1}$, with the gain measured across tasks.
\end{enumerate}
\end{definition}

\paragraph{Relation to continual learning, AutoML, and RSI} Table~\ref{tab:paradigms} compares the two definitions with three neighboring lines of work. Continual learning updates $\Theta$ from a sequence of tasks while limiting forgetting \citep{parisi2019continual,wang2024comprehensive}, including on robots \citep{lesort2020continual,liu2026vla}, and so supplies the retention that physical self-improvement requires. Its task sequence and update rule are fixed by the designer, however, so $\Phi$ does not change. Automated machine learning searches over pipeline, model and training configurations for each problem \citep{hutter2019automated,he2021automl}, which in our terms is mostly a search over the settings of $\Train$. Some systems meta-learn from earlier problems to warm-start later searches \citep{feurer2022autosklearn2}, which updates the procedure in part, but the system found for one problem is not carried over to the next, so changes are not inherited. General RSI frameworks describe an agent by its foundation model, harness, data system, trainer, and improvement mechanism \citep{liu2026path}, which map onto $\Theta$, $H$, $D$, and $\Phi$, and allow all of them to change \citep{tang2026gai,chen2026rsisurvey}. Such frameworks, however, mostly consider digital environments and do not require changes to be physically grounded.

\begin{table}[h]
\centering
\caption{Continual learning, AutoML, and general RSI frameworks compared with the two definitions on six properties. $\checkmark$ typically met, ($\checkmark$) met in part or by some works, $\times$ not met, -- not required or not addressed.}
\label{tab:paradigms}
\small
\setlength{\tabcolsep}{3pt}
\begin{tabular}{@{}>{\raggedright\arraybackslash}p{0.28\textwidth} >{\centering\arraybackslash}p{0.1\textwidth} >{\centering\arraybackslash}p{0.1\textwidth} >{\centering\arraybackslash}p{0.1\textwidth} >{\centering\arraybackslash}p{0.1\textwidth} >{\centering\arraybackslash}p{0.1\textwidth} >{\centering\arraybackslash}p{0.1\textwidth}@{}}
\toprule
Line of work & Made by the system & Inherits changes & Updates $\Theta$, $H$ & Updates $\Phi$ & Recursive gain & Physical grounding \\
\midrule
Continual learning & $\checkmark$ & $\checkmark$ & $\checkmark$ & $\times$ & $\times$ & -- \\
Automated machine learning & $\checkmark$ & $\times$ & $\checkmark$ & ($\checkmark$) & $\times$ & -- \\
General RSI frameworks & $\checkmark$ & $\checkmark$ & $\checkmark$ & $\checkmark$ & $\checkmark$ & -- \\
\midrule
Physical self-improvement & $\checkmark$ & $\checkmark$ & $\checkmark$ & ($\checkmark$) & -- & $\checkmark$ \\
Physical RSI & $\checkmark$ & $\checkmark$ & $\checkmark$ & $\checkmark$ & $\checkmark$ & $\checkmark$ \\
\bottomrule
\end{tabular}
\end{table}

Physical RSI requires real-world grounding, which separates it from RSI in digital environments in four ways.
\begin{enumerate}[leftmargin=*,itemsep=1pt]
\item \emph{Continuous, real-time interaction.} The robot acts through continuous dynamics under real-time constraints, and a task counts as done only when it is completed physically, whereas digital environments can often be paused and replayed.
\item \emph{Costly and irreversible trials.} Resets are expensive and some actions cannot be undone, whereas a digital environment can be restored to an earlier state.
\item \emph{Partial, noisy observation.} The robot sees only local, noisy traces of what happened, whereas digital environments can often expose the full state. Outcomes are therefore uncertain, failures are hard to attribute, and evidence from simulation does not fully carry over to the real robot (the sim-to-real gap).
\item \emph{Safety of people and hardware.} Robots often work alongside people, and the hardware itself can fail or be damaged, so safety constraints, and human oversight where people are involved, need to bound the loops.
\end{enumerate}
These differences are also why our framework requires physical grounding for Physical RSI.

\subsection{System Architecture and Improvement Loops}
\label{sec:architecture}
 
\begin{figure*}[t]
    \centering
    \includegraphics[width=\linewidth]
    {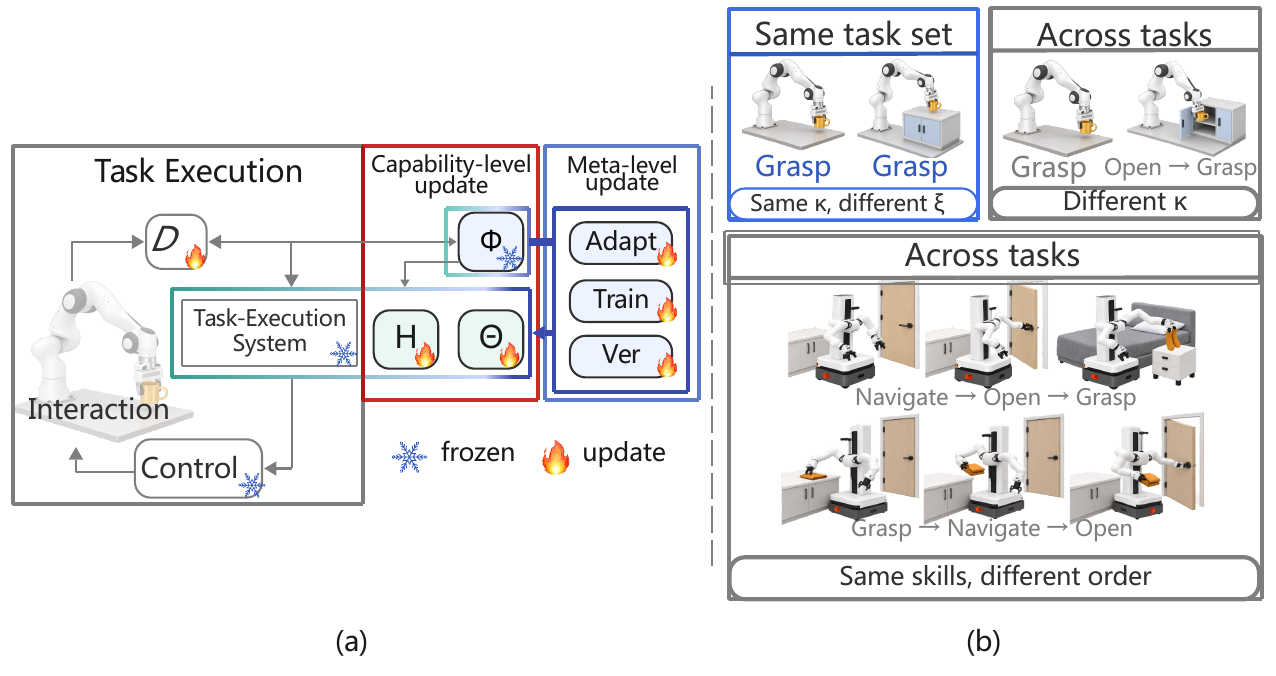}
    \caption{
    Update targets and task identity in the L0--L5 framework.
    (a) Task execution uses the harness $H$ and learned weights $\Theta$. Capability-level updates change $H$ or $\Theta$, whereas meta-level updates change the improvement procedure $\Phi$. $D$ denotes the experience and feedback store.
    (b) Tasks are distinguished by their required skill sequence $\kappa$. Grasping a cup from a table and from a cabinet top belongs to the same task set: $\kappa$ remains unchanged, and only the instance parameters $\xi$ differ. Retrieving a cup from an initially closed cabinet adds opening to $\kappa$ and therefore belongs to a different task set.
    The two long-horizon examples share the same skill set $\kappa_{\mathrm{set}}$ but execute the skills in different orders. Their skill sequences $\kappa$ therefore differ, so they belong to different task sets.
    }
    \label{fig:updates-and-tasks}
\end{figure*}

Section~\ref{sec:definition} defined the elements of an improving robot. This section shows how they work together within a round of Eq.~\eqref{eq:round}. A round runs three loops that differ in what they change (Figure~\ref{fig:updates-and-tasks}(a)), as Table~\ref{tab:processes} summarizes. 

\begin{table}[h]
\centering
\caption{What each loop uses and changes. U, used by the loop; W, written by the loop; --, untouched. The independent evaluator $E$ and the other protected constraints (defined under Safety below) are not listed. $E$ scores the updated task-execution system and judges updates to $\Ver$ from outside.}
\label{tab:processes}
\small
\setlength{\tabcolsep}{3pt}
\begin{tabular}{@{}>{\raggedright\arraybackslash}p{0.24\textwidth} >{\centering\arraybackslash}p{0.065\textwidth} >{\centering\arraybackslash}p{0.065\textwidth} >{\centering\arraybackslash}p{0.065\textwidth} >{\centering\arraybackslash}p{0.065\textwidth} >{\centering\arraybackslash}p{0.065\textwidth} >{\centering\arraybackslash}p{0.065\textwidth} >{\raggedright\arraybackslash}p{0.17\textwidth}@{}}
\toprule
& \multicolumn{2}{>{\centering\arraybackslash}p{0.14\textwidth}}{Task-execution system} & \multicolumn{3}{>{\centering\arraybackslash}p{0.21\textwidth}}{Improvement procedure} & & \\
\cmidrule(lr){2-3}\cmidrule(lr){4-6}
Loop & $H$ & $\Theta$ & $\Adapt$ & $\Train$ & $\Ver$ & $D$ & Effect seen in \\
\midrule
Execution loop & U & U & -- & -- & -- & W & The current episode \\
Capability-level update loop & W & W & U & U & U & U, W & Later tasks \\
Meta-level update loop & U & U & U, W & U, W & U, W & U, W & Changes of later rounds \\
\bottomrule
\end{tabular}
\end{table}

\paragraph{Three loops} In the \emph{execution loop}, the task-execution system uses $H$ and $\Theta$ to turn observations and the episode context into actions, and adds the trajectory to $D$. Behavior that improves within an episode, for example through replanning, ends with the episode and is not evidence of improvement.

In the \emph{capability-level update loop}, $\Phi$ reads $D$ and changes $H$ and $\Theta$, the parts the robot uses when it executes tasks. $\Adapt$ may turn a successful trajectory into a reusable skill or store a lesson from a failure in memory, and $\Train$ may fine-tune a policy on its own successful trajectories or update a world model with observed transitions. The changes that $\Ver$ accepts are inherited, so their effect shows on later tasks, where $E$ measures the gain (Section~\ref{sec:definition}).

In the \emph{meta-level update loop}, $\Phi$ changes one of its own stages, which the robot does not use while executing tasks but which produces or judges its later changes. An updated $\Adapt$ may learn, for instance, to locate the failing module before rewriting a prompt. An updated $\Train$ may switch from imitation learning to reinforcement learning with a reward derived from real outcomes when imitation stops improving a skill. An updated $\Ver$ applies a different standard for accepting candidates, such as other tests, success checks, or thresholds. $\Ver$ can accept an update to $\Adapt$ or $\Train$ by trying the updated stage and judging the changes it produces, whereas an update to $\Ver$ is judged by $E$. 

\paragraph{Two pathways: $H$ with $\Adapt$, $\Theta$ with $\Train$} Updates follow two pathways, each pairing a part of the task-execution system with the stage of $\Phi$ that updates it. The \emph{harness pathway} pairs $H$ with $\Adapt$ and is reviewed in Section~\ref{sec:harness}, and the \emph{training pathway} pairs $\Theta$ with $\Train$ and is reviewed in Section~\ref{sec:training}. Both run through the capability level and the meta level. The pathways interact. Skills and recovery routines added to $H$ change which trajectories enter $D$ for training, and an updated $\Theta$ changes which harness edits are worth proposing. Updates to $\Ver$ lie on neither pathway and are covered under evaluation (Section~\ref{sec:evaluation}).

\paragraph{Feedback and safety} Feedback is the information that reaches later updates through $D$. It comes from task execution, which supplies outcomes, internal signals, and human corrections, and from updates, which leave a record of what was changed and how $\Ver$ or $E$ judged it. Errors in feedback travel the same route. A misdetected success can lead $\Ver$ to accept a harmful change, and a misattributed failure can send $\Adapt$ to edit the wrong module.
For safety, it constrains all three loops rather than being checked once after training, because physical trials can be irreversible and robots often share space with people (Section~\ref{sec:definition}). During execution, action filters and human takeover keep trials safe, and $\Ver$ rejects updates that violate safety requirements or degrade earlier capability. Since an updated $\Ver$ might relax these tests, the safety rules are kept out of reach of $\Phi$ and, together with $E$, form the \emph{protected constraints} (Section~\ref{sec:safety}).

\subsection{The L0--L5 Framework}
\label{sec:levels}

Figure~\ref{fig:overview}(b) and Section~\ref{sec:intro} introduced the levels. This section states when a workflow meets each of them. The answer depends on what is changed, what counts as a task, and how far the gain extends beyond it. Figure~\ref{fig:updates-and-tasks} illustrates the update targets and the distinction between same-task and across-task evaluation.

\paragraph{Where a component sits} We assign each component to $H$, $\Theta$, or $\Phi$ by whether the task-execution system uses it. A component used to carry out a task belongs to $H$ or $\Theta$. LRLL \citep{tziafas2024lrll} keeps a library of executable skills that the policy code calls during task execution, so the library is part of $H$. ASPIRE \citep{lu2026aspire} keeps a library of guidance that the coding agent reads when it writes the program for a new task, so this library is part of $H$ as well. VERITAS \citep{zhang2026veritas} uses a VLM to generate a visual trace and a geometric verifier to score candidate actions during execution. The VLM weights belong to $\Theta$, whereas the trace representation and geometric scoring logic belong to $H$.

A component used only to produce or judge retained changes belongs to $\Phi$. This covers procedures that run only during improvement, such as the reward code of Eureka \citep{ma2024eureka}, the training and algorithm code of ENPIRE \citep{xiao2026enpire}, and the diagnosis, admission, and search procedures with which ASPIRE builds its library. It also covers learned components that only improvement uses. One kind is the reward model or success detector, as in the reward classifier of MEDAL++ \citep{sharma2023medalpp} and the success classifiers of HIL-SERL \citep{luo2024hilserl} and ConRFT \citep{chen2025conrft}. Another is a world model that only serves as the training environment, as in RISE \citep{yang2026rise}, whereas a world model used for planning during task execution belongs to $\Theta$. Changes to $H$ and $\Theta$ are capability-level updates. But not every change to a component in $\Phi$ constitutes a meta-level update. During online policy training, critics and reward classifiers may be repeatedly fitted according to a fixed training procedure. Their estimates change as training proceeds, but the procedure itself remains unchanged. These updates are part of running the existing improvement mechanism. A meta-level update changes the mechanism itself, such as its objective, update rule, training configuration, curriculum, or selection criterion, and retains that change for subsequent improvement. Such a change may be implemented in code or learned parameters.

\paragraph{Tasks} A task is what the experimenter asks the robot to accomplish, such as lifting a cup or placing it in the sink. We describe it by two factors. The skill sequence $\kappa$ is the sequence of skills required to accomplish the task, such as walking to an object, opening a door, grasping, and pouring. The instance parameters $\xi$ are the variation that leaves $\kappa$ unchanged, such as the poses of objects, their arrangement, and the difficulty of a scene as long as the same skills still suffice. This follows the distinction in Meta-World \citep{yu2019metaworld}. Whether the scene is simulated or real is recorded as the source of evidence and does not change the task.

Whether a task has changed depends on its $\kappa$, which can change in two ways (Figure~\ref{fig:updates-and-tasks}(b)). The first is that $\kappa$ gains a new skill. Lifting a cup that stands in front of the robot and walking to a table before lifting it are different tasks, because only the second requires navigation. Walking to a table before lifting a cup and walking to a kitchen counter before lifting it are the same task, because both require only navigation and grasping. The second is that $\kappa$ is a new combination of skills that the system already has, which is what a long-horizon task usually is. Going from the kitchen to the bedroom and fetching a pair of socks runs navigation, opening the door, and grasping, whereas picking up a piece of clothing in the kitchen and carrying it into the bedroom runs grasping, navigation, and opening the door. The skills are the same but the order differs, so the two are different tasks. This matches the hardest level of GemBench \citep{garcia2025gembench}, where the robot must compose several skills learned during training.

We apply this distinction to evaluation protocols from seven surveyed systems. Five of these protocols do not establish across-task improvement. HIL-SERL \citep{luo2025hilserl} trains and tests a policy on USB grasp-insertion, with variation in the initial USB placement. The required skill sequence remains grasping followed by insertion, so only $\xi$ changes. In the pin-insertion experiment considered here, ENPIRE \citep{xiao2026enpire} trains and evaluates on the same task. ReWiND \citep{zhang2025rewind} fine-tunes on a task variation new to the pretrained model but evaluates on that same variation. Eureka \citep{ma2024eureka} separately optimizes a reward for each of 29 tasks and evaluates the resulting policy on the corresponding task. In these four cases, improvement and evaluation concern the same $\kappa$. Eurekaverse \citep{liang2024eurekaverse} trains on diverse generated parkour courses and evaluates on held-out courses, including combinations of familiar skills. However, the reported protocol does not establish whether the required skill sequences in these test courses were already covered during training. We therefore cannot determine whether the tests change only $\xi$ or also $\kappa$, and conservatively do not treat these results as evidence of across-task improvement. The other two are evaluated across tasks, because $\kappa$ changes. ASPIRE \citep{lu2026aspire} constructs its skill library from LIBERO-90 \citep{liu2023libero}, whose tasks are short-horizon, and applies the library zero-shot to held-out long-horizon tasks from LIBERO-Pro Long, which require composing these skills into longer sequences. LRLL \citep{tziafas2024lrll} is tested on unseen compositions of tasks from the current and earlier learning cycles. In both, the test tasks are new combinations of skills the system already has, which is the second way in which $\kappa$ can change.

\begin{figure*}[h]
    \centering
    \includegraphics[width=\textwidth]{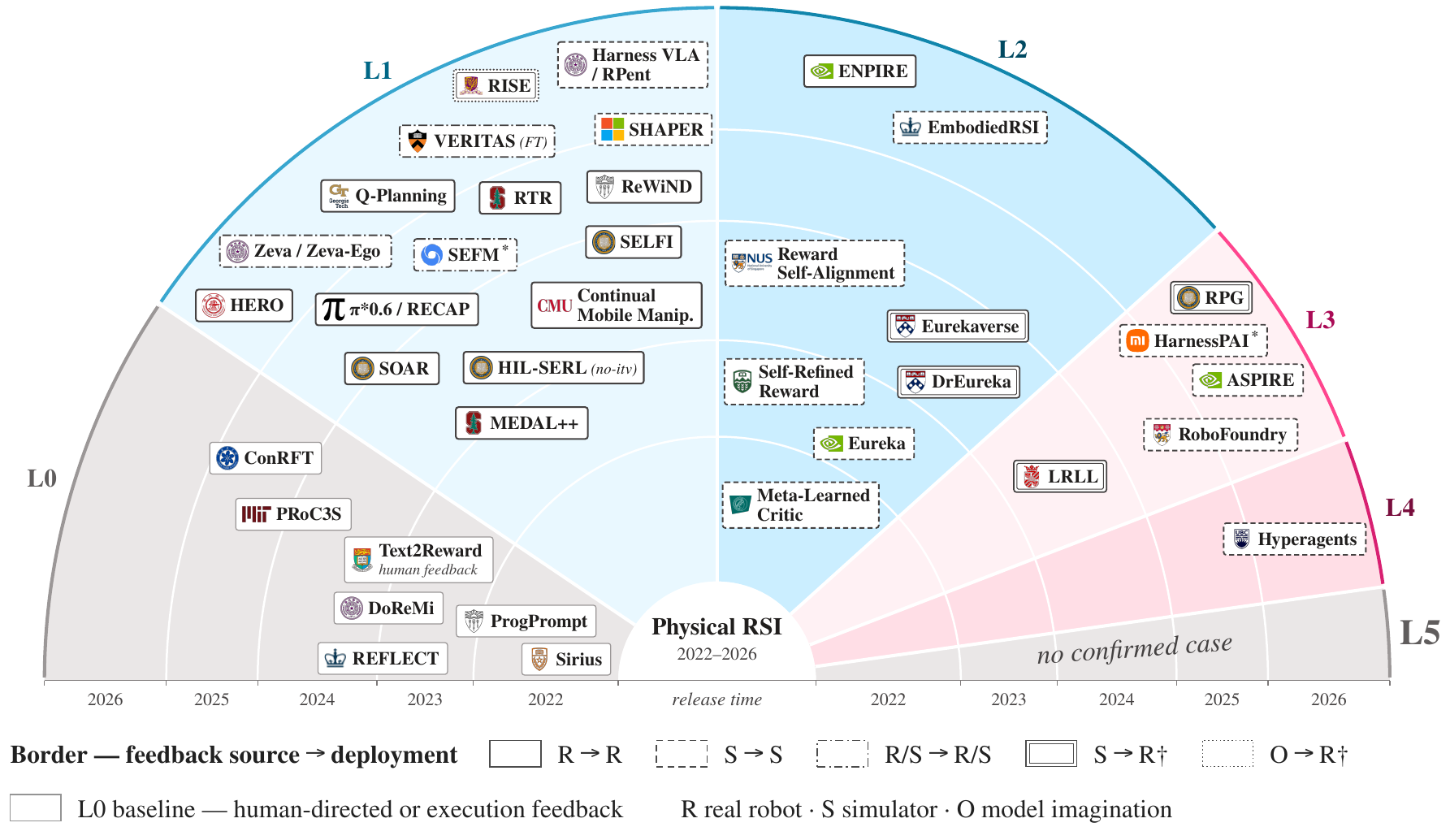}
    \caption{Selected systems by level, year, and evidence source. Each system sits in the sector of its level, and its distance from the centre gives its release year, from 2022 near the centre to 2026 at the rim. The border of each card gives the source of the feedback behind the update followed by the setting of deployment. R, real; S, simulation; O, model imagination, meaning a learned world model. A dagger marks a deployment that gave no feedback to the marked improvement loop: the update is driven by simulated (S) or imagined (O) feedback and the system is then deployed or validated on a real robot, which does not mean that the paper used no real data. L0 cards form a baseline of systems directed by humans or by execution feedback, and parenthetical labels name the branch evaluated. An asterisk is an affiliation note: SEFM* shows the affiliation at the time of the work rather than the first author's current one, and HarnessPAI*, which is credited to a team, shows the affiliation of its first core contributor.}
    \label{fig:evidence-map}
\end{figure*}
 
\paragraph{Evaluation scope and evidence sources}
A system is scored on the independent evaluator $E$ of Section~\ref{sec:definition}, and which tasks of $E$ count depends on the claim. A same-task claim is scored on the tasks of $E$ that have the same $\kappa$ as the tasks used for improvement, and an across-task claim on the tasks of $E$ with a different $\kappa$. An across-task claim also requires that a change retained from improvement on the source tasks is used on the target tasks and contributes to their performance. We write $J_Q^{\sigma}$ for the resulting score under $Q$, with $\sigma$ either same task or across tasks. Section~\ref{sec:evaluation} describes this measurement in detail. For each system, we also record the source of the feedback driving the update and the setting in which the updated system is deployed, each as real, simulated, or imagined.  
 
\paragraph{Levels} We assign a level by four questions. The first is whether $\Phi$ produces a retained change from the system's own experience. Changes inside a task execution are not improvement (Section~\ref{sec:architecture}), so a workflow with no such retained change is L0. We also place at L0, as a reference group, workflows whose retained changes depend on people intervening or labeling during improvement, so that L1 to L5 credit only autonomous improvement. The second is what the retained change updates, $H$ or $\Theta$ at the capability level, or a stage of $\Phi$ at the meta level (Section~\ref{sec:architecture}). The third is the scope of the gain, which is measured on the same task or across tasks, judged by $\kappa$ as above. The fourth is whether the retained changes give a gain, meaning that the final system scores higher than the initial one on $J_Q^{\sigma}$. A workflow with a gain is placed by its second and third answers, giving L1 to L4 as in Figure~\ref{fig:overview}(b). L5 is an L4 workflow that also shows recursive gain. Starting from the same generation and under the same budget, the system produced by each updated $\Phi_k$ scores higher than the one produced by $\Phi_{k-1}$, and the changes it relies on are physically grounded. Figure~\ref{fig:evidence-map} organizes selected systems by these levels, together with their year, update feedback source, and deployment setting.

\section{Harness Adaptation of Physical RSI}
\label{sec:harness}
\label{sec:agent}

\begingroup
\setlength{\intextsep}{6pt}
\begin{figure}[htb]
\setlength{\abovecaptionskip}{3pt}
\setlength{\belowcaptionskip}{0pt}
\centering
\includegraphics[width=\linewidth]{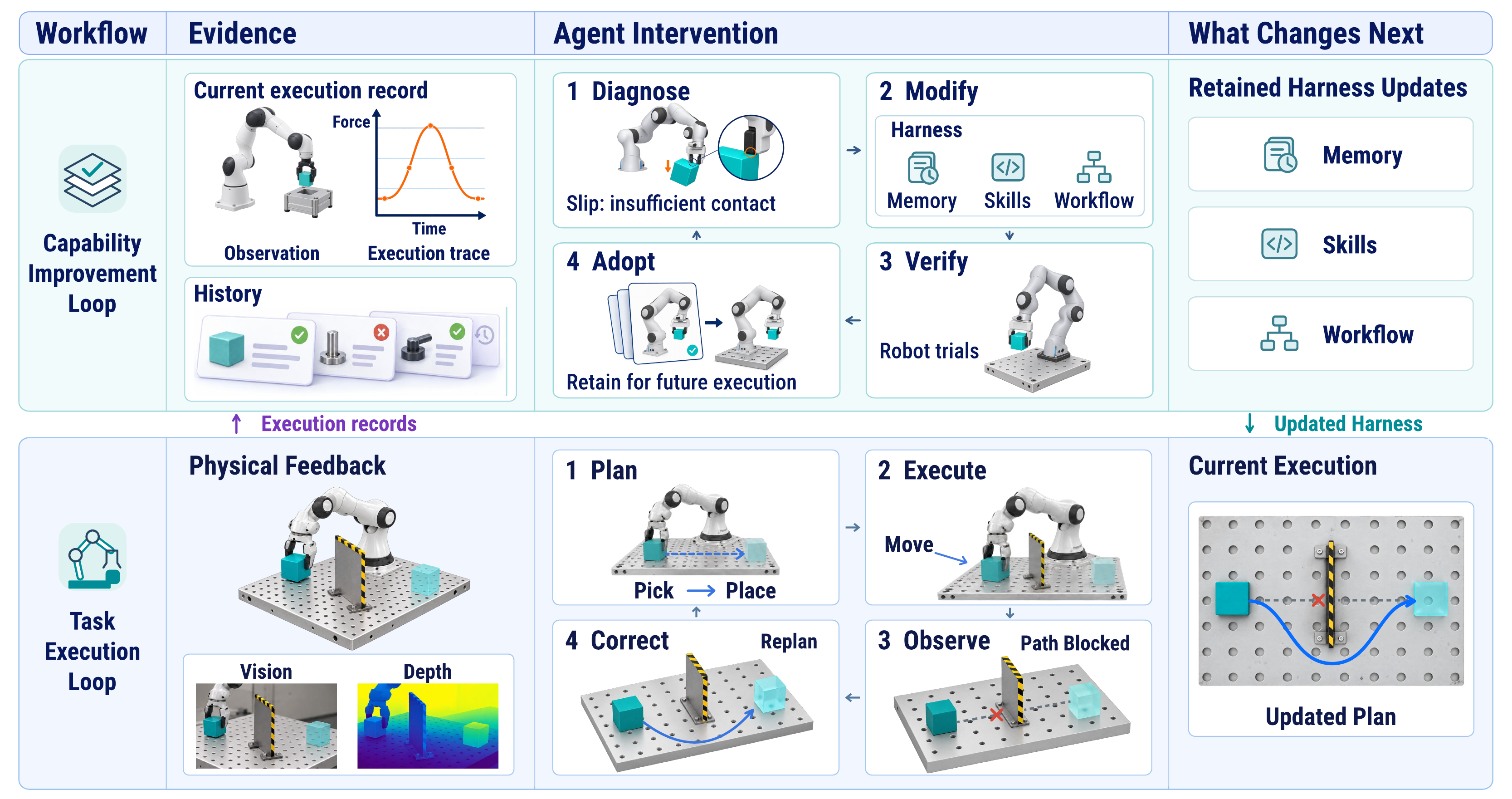}
\caption{Task execution and capability improvement. Execution records inform persistent updates, and adopted harness updates support subsequent execution. The harness pathway changes task resources and execution procedures.}
\label{fig:harness-execution-improvement}
\end{figure}
\endgroup

The task execution loop connects planning, execution, observation, and correction (Figure.~\ref{fig:harness-execution-improvement}). Robot skills and controllers carry out planned actions, while physical feedback informs adjustments to the current behavior or remaining plan. The harness supports this loop through memory, executable skills, and workflows.

Memory supplies decision context through environmental representations and retained experience, as illustrated by object localization in DynaMem~\citep{liu2024dynamem} and the retrieval of corrective guidance in DROC~\citep{agentdraft_droc}. Skills expose robot capabilities through callable programs and interfaces. Code as Policies~\citep{liang2022codeaspolicies} composes perception outputs and control primitives in executable code, while SkiROS2~\citep{mayr2023skiros2} organizes parameterized skills with explicit execution conditions, exposing program logic, skill parameters, and interfaces as potential modification targets. Workflows govern capability selection and coordination through mechanisms such as feasibility-based selection in SayCan~\citep{ahn2022icanisay} and sequencing and fallback in behavior trees~\citep{colledanchise2018behaviortrees}. Together, these components support adjustments to the current execution, including revised action choices, corrected motion, and updated plans.

The capability improvement loop uses execution feedback and prior modification history to produce persistent updates to memory, skills, and workflows. Feedback interpretation identifies modification targets and scope, guiding the construction of candidate changes. Adopted updates become available for subsequent execution, while attempted changes and their outcomes accumulate as improvement history to inform future updates.

\subsection{Feedback Interpretation and Update Target Identification}
\label{sec:harness-feedback}

Execution records and improvement histories support diagnosis by relating observed deviations to the actions, conditions, and previous interventions associated with them. In simulated robot tasks, primitive-level traces in ASPIRE~\citep{lu2026aspire}, workflow-node diagnostics in HarnessPAI~\citep{agentdraft_hsa_harnesspai2026}, and visual keyframes in Act-Observe-Rewrite~\citep{agentdraft_kumar2026actobserverewritemultimodalcodingagents} connect execution outcomes to specific operations and observations. Comparisons across attempts provide additional grounds for interpreting these deviations. In digital agent studies, ExpeL~\citep{agentdraft_zhao2024expel} contrasts successful and failed attempts, while AutoGuide~\citep{agentdraft_hsa_autoguide} relates their differences to the contexts in which corrective guidance applies. Modification histories in WikiSkill~\citep{agentdraft_hsa_wikiskill} and language corrections in DROC~\citep{agentdraft_droc} supplement these comparisons with evidence from previous edits and clarification of task requirements. Together, these sources support hypotheses about the causes of observed failures and the conditions under which they recur.

Selecting a modification target requires determining which harness component should change in response to the diagnosis. When failures concern the use of guidance, AutoManual~\citep{agentdraft_hsa_automanual} and EmbodiSkill~\citep{agentdraft_hsa_embodiskill2026} distinguish defective instructions from failures to follow valid instructions, directing intervention toward the instructions themselves or reminders of their use. Dynamo~\citep{agentdraft_hsa_dynamo} distinguishes reasoning from perceptual gaps to target textual skills or perception tools. These judgments must then be grounded in editable implementations. HarnessFix~\citep{agentdraft_hsa_harnessfix} links failed trajectory steps to harness code, while RACaP~\citep{agentdraft_hsa_racap2026} identifies interventions involving interfaces, exposed parameters, reporting, routing, or memory. Once the component is identified, the proposal must delimit both the extent of the change and its intended applicability. Self-Harness~\citep{agentdraft_hsa_self_harness} favors bounded changes, and SkillAdam~\citep{agentdraft_hsa_skilladam} uses prior editing outcomes to regulate their scope, while RoboFoundry~\citep{agentdraft_liang2026robofoundry} distinguishes task-specific repairs from changes intended for broader reuse. With the diagnosed problem linked to a concrete implementation and a defined scope, the next step is to construct a modification that addresses it within these bounds.

\begingroup
\setlength{\intextsep}{6pt}
\begin{figure}[htb]
\setlength{\abovecaptionskip}{3pt}
\setlength{\belowcaptionskip}{0pt}
\centering
\includegraphics[width=\linewidth]{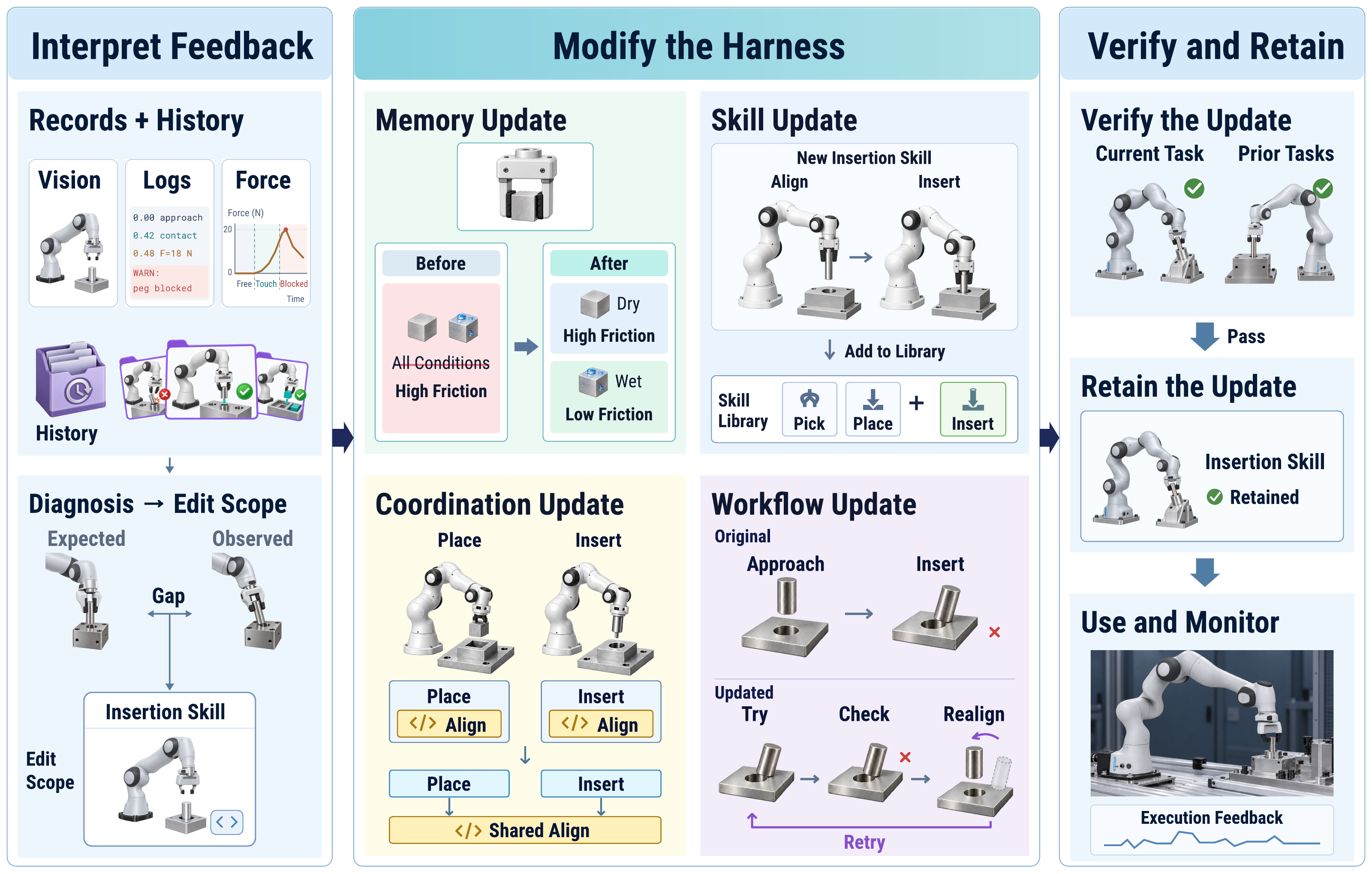}
\caption{Harness adaptation from execution feedback to retained updates. Diagnosis identifies edit targets, memory, skill, and workflow changes modify the harness, and verification and subsequent execution inform adoption and continued use.}
\label{fig:harness-adaptation}
\end{figure}
\endgroup

\subsection{Memory, Skill, and Workflow Adaptation}
\label{sec:harness-skills}

Execution memory updates add experiential guidance, organize its application, and revise outdated content (Figure.~\ref{fig:harness-adaptation}). In digital agent tasks, Reflexion~\citep{agentdraft_shinn2023reflexion} retains feedback-derived reflections, while ExpeL~\citep{agentdraft_zhao2024expel} extracts reusable insights across attempts. These additions can be organized into context-specific guidance, as in AutoGuide~\citep{agentdraft_hsa_autoguide}, or maintained as editable rule collections, as in AutoManual~\citep{agentdraft_hsa_automanual}. Related execution guidance can also be maintained through tool documentation. ContDA~\citep{agentdraft_hsa_wu2026contda} revises tool documentation, usage preferences, and fallback guidance as toolsets evolve. In robotics, memory revision additionally concerns the physical explanations underlying action selection. PhysMem~\citep{agentdraft_hsa_physmem_2026} develops interaction-derived hypotheses and tests their applicability, while KnowBody~\citep{agentdraft_lou2026knowbody} revises action-relevant body relations and rechecks dependent knowledge. Together, these mechanisms update what subsequent execution can draw on and the conditions under which it applies.

Individual capabilities can be updated through changes to their implementation or interface, or supplemented by newly constructed skills and tools. CREATOR~\citep{agentdraft_hsa_qian2023creator} uses execution errors to guide tool-code correction, while ASPIRE~\citep{lu2026aspire} repairs robot task programs using feedback associated with primitive calls. Act-Observe-Rewrite~\citep{agentdraft_kumar2026actobserverewritemultimodalcodingagents} generates replacement perception and controller code between simulated manipulation trials, changing the computations that interpret observations and produce actions. Other adaptations concern how an existing capability is exposed. RACaP~\citep{agentdraft_hsa_racap2026} includes API and parameter-exposure changes among its candidate modifications. New capabilities can instead be constructed directly or abstracted from task solutions. LATM~\citep{agentdraft_hsa_cai2023latm} separates tool construction from subsequent invocation, allowing generated functions to serve repeated requests. ViReSkill~\citep{agentdraft_kagaya2025vireskill} stores successful robot plans and control code for later reuse. CRAFT~\citep{agentdraft_hsa_yuan2024craft} and SkillWeaver~\citep{agentdraft_hsa_zheng2025skillweaver} construct parameterized tools and reusable web APIs, respectively, while Playful Agentic Robot Learning~\citep{agentdraft_zhang2026playful} extracts reusable helpers from successful robot programs in simulation. Such changes concern individual capabilities, but shared implementations and callers can require coordinated modification.

Coordinated updates change related capabilities together when they share functionality or depend on common interfaces. ReGAL~\citep{agentdraft_hsa_regal} couples the introduction of helper functions with refactoring of the programs that use them, linking library construction to changes in callers. In simulation with human demonstrations and hints, Lifelong Robot Library Learning~\citep{tziafas2024lrll} groups robot programs by structural similarity, extracts parameterized skills, and rewrites existing examples to use the resulting library. In both cases, introducing an abstraction is accompanied by changes to existing programs that previously implemented the functionality separately. Coordination also concerns the maintenance of these shared structures. ReGAL edits or prunes abstractions that perform poorly on their associated tests. Playful~\citep{agentdraft_zhang2026playful} extends library-level modification through periodic consolidation of redundant helpers, with dependencies recorded in library entries. The relevant unit of change therefore includes shared functionality and its dependent programs, rather than a collection of independently modified skills.

Execution workflows govern when these capabilities are invoked and how their outputs advance or redirect the task. Harness VLA~\citep{agentdraft_hsa_harnessvla2026} explores staging, action ordering, and invocation timing, retaining successful execution experience to guide later attempts. BATON~\citep{agentdraft_hsa_baton2026} makes the conditions connecting successive capabilities explicit through invocation, handoff, and lookahead memories. These records guide when a frozen policy can be invoked, what entry state a subsequent subtask requires, and which execution strategy leaves a suitable state for its successor. Changes can also affect the mechanisms that track and advance task progress. RoboFoundry~\citep{agentdraft_liang2026robofoundry} persistently modifies stage processing, active-subgoal construction, and replanning cadence, allowing subsequent executions to use updated procedures for continuing unfinished work. In simulation, RACaP~\citep{agentdraft_hsa_racap2026} evolves its ReAct harness before deployment, then freezes the implementation while retaining observation-dependent action selection, recovery, and stopping decisions. Recovery procedures can likewise be represented as explicit editable rules. In text-agent and simulated object-centric tasks, Kintsugi~\citep{agentdraft_hsa_kintsugi2026} updates policy schemas, monitors, and recovery entries between rollouts. In digital interaction tasks, Living-Harness~\citep{agentdraft_hsa_living_harness_2026} converts completed-episode feedback into persistent trigger--recovery records and state-conditioned transitions. Across these representations, updates alter capability invocation, task progression, and recovery in subsequent executions.

\subsection{Adoption and Improvement History}
\label{sec:harness-verification}
\label{sec:harness-adoption}

Candidate harness updates are checked for their intended effects and their impact on existing capabilities. ASPIRE~\citep{lu2026aspire} validates repairs through re-execution, while ReGAL~\citep{agentdraft_hsa_regal} uses tests of affected programs to assess changes to shared functions and their callers. These checks inform candidate acceptance, with verification methods discussed further in Section~\ref{sec:evaluation-verification}. Accepted updates become resources available to subsequent execution, including reusable robot plans and control code in ViReSkill~\citep{agentdraft_kagaya2025vireskill} and persistent library helpers in Playful Agentic Robot Learning~\citep{agentdraft_zhang2026playful}. Continued execution provides evidence for reassessing adopted resources. In PhysMem~\citep{agentdraft_hsa_physmem_2026}, subsequent interactions strengthen or weaken confidence in retained explanations. In KnowBody~\citep{agentdraft_lou2026knowbody}, changes to body estimates prompt renewed checks of dependent rules, linking continued use to the validity of their underlying assumptions. Adoption thus connects candidate updates to subsequent execution, whose feedback informs continued retention, renewed verification, or further modification.

\begingroup
\setlength{\intextsep}{6pt}
\begin{figure}[htb]
\setlength{\abovecaptionskip}{3pt}
\setlength{\belowcaptionskip}{0pt}
\centering
\includegraphics[width=\linewidth]{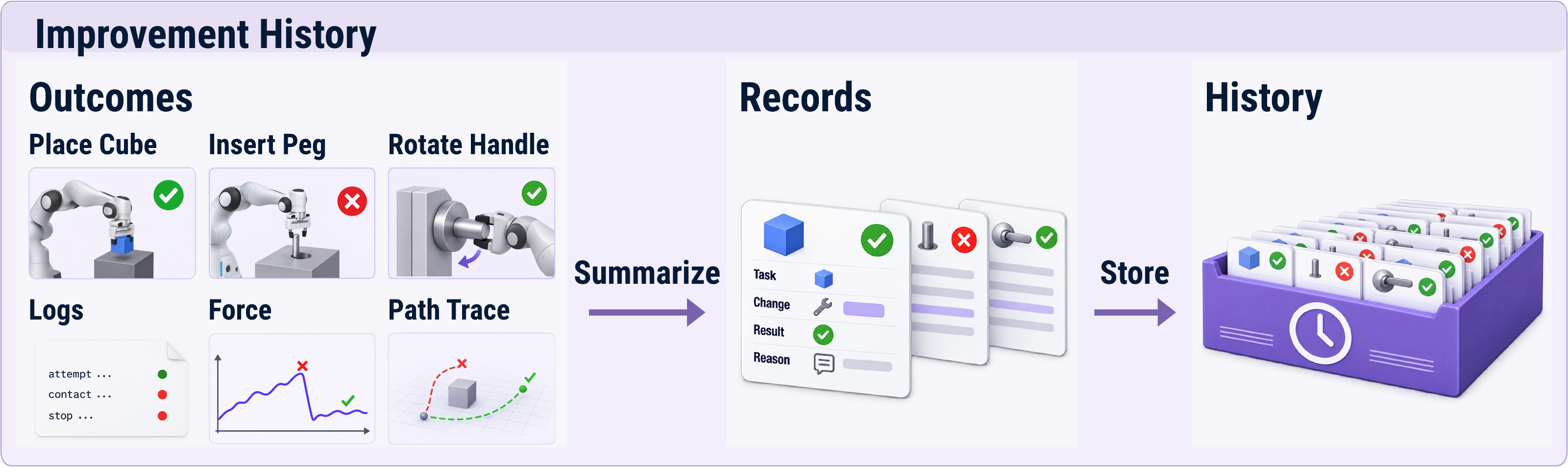}
\caption{Improvement history. Outcomes and execution evidence from different tasks are summarized into modification records and stored for later diagnosis and improvement. Successful and unsuccessful attempts both contribute to the history.}
\label{fig:harness-repair-history}
\end{figure}
\endgroup

Improvement history links modification attempts to their execution evidence, outcomes, and supporting rationale (Figure~\ref{fig:harness-repair-history}). ASPIRE~\citep{lu2026aspire} records experiments, attempted fixes, skill updates, and observed results, while KnowBody~\citep{agentdraft_lou2026knowbody} preserves original observations and actions alongside outcome assessments, allowing later interpretations to remain traceable to physical evidence. Structured records connect this evidence to specific changes. In digital and simulated tasks, WikiSkill~\citep{agentdraft_hsa_wikiskill} records target skills, edit differences, validation scores, and acceptance decisions, while EvoSkill~\citep{alzubi2026evoskill} accumulates proposals, their rationale, and performance changes, retaining feedback even when candidates are discarded. Across attempts, SkillAdam~\citep{agentdraft_hsa_skilladam} organizes modifications and outcomes by issue, preserving resolution histories and reopening issues when failures recur. Records can also associate update methods with their products, as in MCE~\citep{agentdraft_hsa_mce2026}, which stores context-engineering strategies, generated functions, and evaluation metrics together. These practices turn individual attempts into a persistent history for subsequent diagnosis and proposal selection, while making the methods behind those attempts available for further improvement.

\subsection{Meta-Improvement of Harness Adaptation}
\label{sec:harness-improvement}

\begingroup
\setlength{\intextsep}{6pt}
\begin{figure}[htb]
\setlength{\abovecaptionskip}{3pt}
\setlength{\belowcaptionskip}{0pt}
\centering
\includegraphics[width=\linewidth]{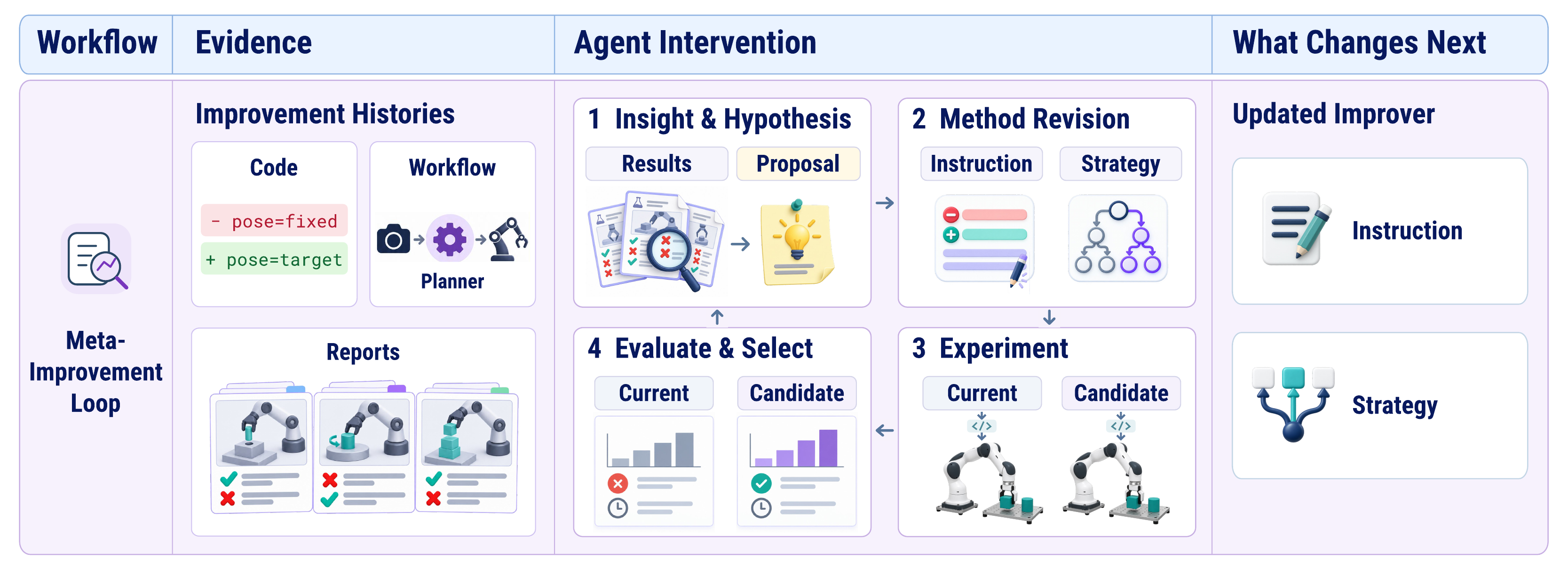}
\caption{The meta-improvement loop. Improvement histories inform hypotheses and method changes. Current and candidate adaptation methods are compared through the subsequent harness updates they produce under matched budgets. Evaluation of subsequent gains and costs supports adoption, further revision, or rejection of the candidate method. Adopted methods govern future improvement.}
\label{fig:harness-meta-learning}
\end{figure}
\endgroup

Meta-improvement modifies \(\mathrm{Adapt}\), the method that generates subsequent harness updates, using evidence accumulated across improvement attempts. Capability improvement can change retained resources and execution procedures under a fixed adaptation method, whereas meta-improvement changes the method itself (Figure~\ref{fig:harness-meta-learning}). G\"odel Machines~\citep{agentdraft_schmidhuber2003goedel} establish a formal precedent for self-modification, while STOP~\citep{zelikman2023stop} and Hyperagents~\citep{prsi_hyperagents} implement editable improvement procedures in digital systems. For Physical RSI, experiment latency and resource constraints, discussed more broadly in \emph{From AGI to ASI}~\citep{agentdraft_genewein2026agiasi}, motivate improving the quality of updates produced within limited interaction budgets.

Editing experience, difficult cases, and prior evaluations provide different grounds for revising the guidance, resource-construction operations, and programs used by \(\mathrm{Adapt}\). At the guidance level, SkillOpt~\citep{agentdraft_hsa_skillopt2026} consolidates editing experience into an optimizer-side meta skill, while Promptbreeder~\citep{agentdraft_hsa_fernando2023promptbreeder} evolves mutation prompts according to the performance of the task prompts they generate. Difficult cases can expose limitations in resource construction, while evaluations of previous designs guide the search for alternatives. MemSkill~\citep{agentdraft_hsa_memskill2026} uses difficult cases to refine or extend memory-construction skills, MCE~\citep{agentdraft_hsa_mce2026} evolves context-building strategies from previous strategies and evaluations, and ALMA~\citep{agentdraft_hsa_alma2026} searches designs containing memory-update functions using prior designs and interaction results. These approaches modify the operations constructing retained context or memory rather than merely adding stored content. At the program level, VERSE~\citep{agentdraft_hsa_verse2026} modifies optimizer-side diagnostic tools and calls to existing checks, while STOP~\citep{zelikman2023stop} and Hyperagents~\citep{prsi_hyperagents} make the update-generating procedure itself editable.

The value of a modified adaptation method depends on the effectiveness and cost of the subsequent updates it produces. STOP~\citep{zelikman2023stop} evaluates improvers through the downstream utility of generated programs, while Hyperagents~\citep{prsi_hyperagents} and SelfSearch~\citep{agentdraft_hsa_selfsearch} use fixed-improver comparisons to examine the contribution of method evolution. Attribution becomes less direct when adaptation methods and runtime components change together, as in MemEvolve~\citep{agentdraft_zhang2025memevolve} and ALMA~\citep{agentdraft_hsa_alma2026}, and VERSE~\citep{agentdraft_hsa_verse2026} reports that method evolution does not yield gains in every setting. Evaluation must also distinguish useful improvements from exploitation of the measurement process. \citet{huang2026rewardhacking} document this risk in digital research tasks, motivating independently checked evidence when evaluating autonomous method changes. For Physical RSI, comparisons under matched initial systems, experience, and budgets should establish whether a candidate method improves later adaptation, with detailed protocols addressed in Section~\ref{sec:evaluation-outcomes}.

Evaluation outcomes determine whether a candidate method is adopted, revised, or rejected. Adoption requires the changed procedure to govern subsequent improvement attempts. HSI~\citep{agentdraft_hsa_hsi2026} explicitly executes revised evolution strategies in later rounds, while VERSE~\citep{agentdraft_hsa_verse2026} conditions optimizer replacement on code checks and a test run, retaining the previous implementation when these checks fail. In Physical RSI, the inherited \(\mathrm{Adapt}\) must produce harness updates that undergo verification and subsequent physical execution. Their outcomes extend the improvement history, providing evidence for retaining the method or targeting further changes to its guidance, operations, or program. This closes the method-level feedback loop, although recursive gains require evidence that successive inherited revisions improve later adaptation, as specified in Definition~\ref{def:prsi}.

\section{Model Training of Physical RSI}
\label{sec:training}

\begin{figure}
    \centering
    \includegraphics[width=1\linewidth]{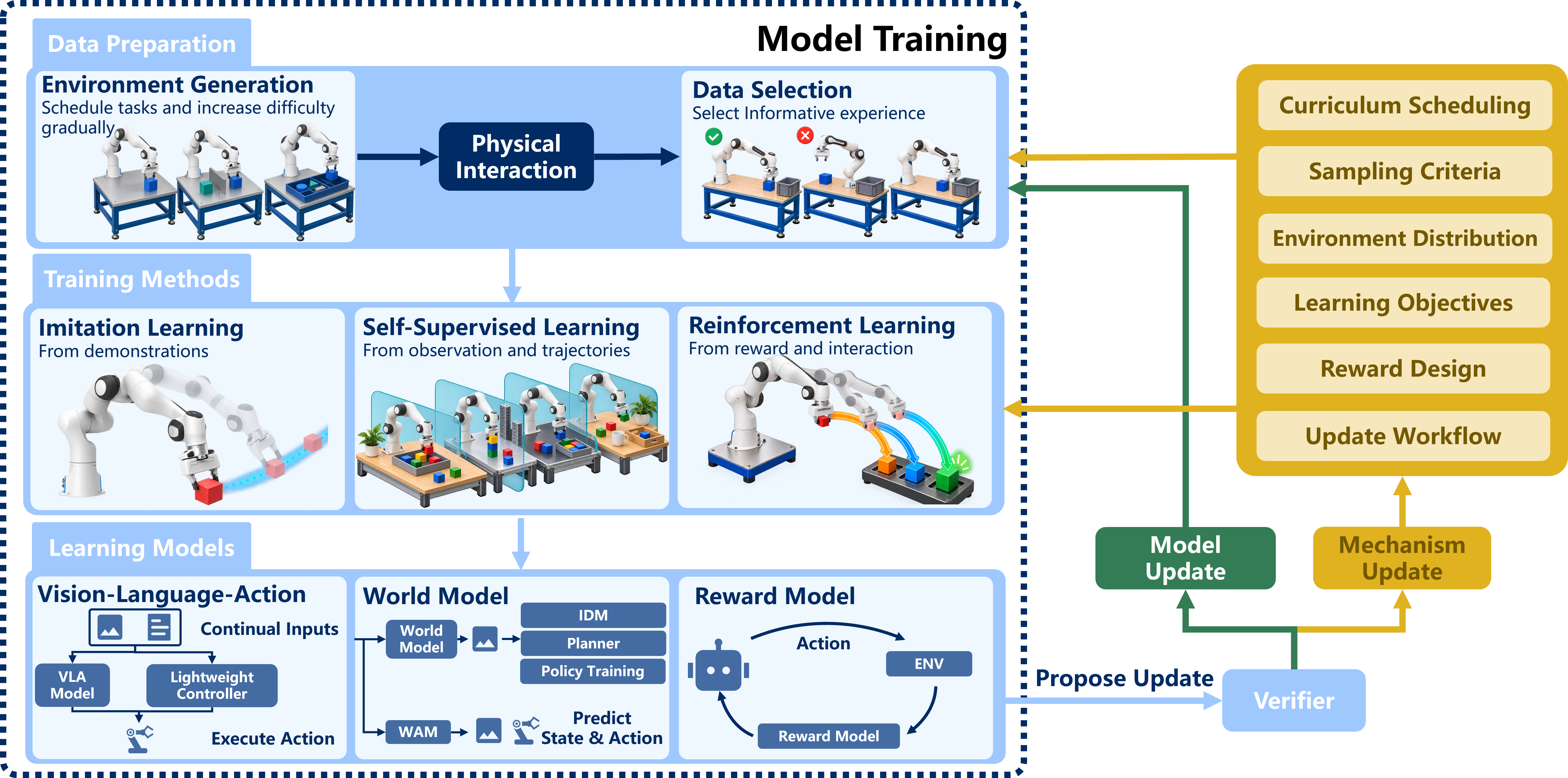}
    \caption{Model-learning workflow in physical RSI. Environment generation and data selection organize experience from physical interactions for training policies, world models, and reward models through imitation, self-supervised, or reinforcement learning. Candidate updates undergo verification before adoption: model updates shape subsequent execution and learning, while mechanism updates revise how data are prepared and models are trained in later rounds.}    
    \label{fig:model-learning-workflow}
\end{figure}

As illustrated in Figure~\ref{fig:overview}, model learning provides another pathway for improvement alongside adaptation of the model-external harness. Physical experience can support updates to policies, reward models, and world models. A frozen foundation model may assist with failure analysis and training decisions. Figure~\ref{fig:model-learning-workflow} zooms into this pathway and shows how model learning can form a continuous improvement loop: Physical interaction provides experience for training, and the resulting model updates shape subsequent interactions, forming a feedback loop that may eventually extend to the training procedure itself. This distinction between improving a learnable model and modifying the process that produces future model updates parallels the L0--L5 distinction between task-level improvement and adaptation of the improvement mechanism. Section~\ref{sec:physical-rsi-evaluation} examines the reliability of the feedback on which both pathways depend.

\subsection{Data Preparation}
\label{sec:training-data-preparation}

Data preparation couples experience generation with selection for learning. Physical practice prioritizes weak capabilities under scene and reset constraints~\citep{sharma2023selfimprovingrobots,fang2026find}. Simulation generators construct tasks, scenes, rewards, and demonstrations~\citep{wang2023gensim,hua2024gensim2scalingrobotdata,curricullm,wang2024robogenunleashinginfinitedata}, expanding coverage and adjusting difficulty to learner performance~\citep{choi2026generative3d,kang2026simworldstudio}. Real2Sim2Real workflows further connect environment preparation to deployment by reconstructing observed scenes in simulation and using the resulting spatial context for physical execution. HomeBody illustrates this route: a foundation-model agent builds a digital twin from robot observations and SLAM geometry to support reasoning and skill execution~\citep{huh2026homebody}. Selection determines how this experience is used. Shared control, active querying, and correction memory provide human guidance~\citep{liu2023sirius,luo2024hilserl,hoque2021thriftydagger,zero2skill}, while recovery demonstrations cover failure states~\citep{ge2026egorecovery}; dependence on essential human corrective judgments places these workflows at L0. Autonomous selection admits verified successes for imitation and uses failures for value learning or recovery~\citep{soar,hao2026far}; prioritizing failure-prone states focuses scarce trials on weak behavior~\citep{he2026criticality}. Data preparation contributes to mechanism improvement when a data sampler, curriculum generator, or selection model is revised and reused in subsequent learning~\citep{chebotar2019simopt,muratore2021bayrn,muratore2022npdr,sekar2020plan2explore}.

\subsection{Training Methods}
\label{sec:training-method}

Model training typically begins with pretraining on diverse data to establish behavioral, visual, or dynamics priors, followed by post-training with task-specific experience or feedback~\citep{kawaharazuka2025vlasurvey}. Across pretraining and post-training, we review three learning paradigms: imitation learning (IL), self-supervised learning (SSL), and reinforcement learning (RL).

\textbf{Imitation learning} (IL) learns behavior from demonstrations or teacher-generated targets; selected self-generated experience can also supervise later updates~\citep{bousmalis2024robocat}. Behavior cloning can fit continuous actions by regression or predict discrete action tokens autoregressively~\citep{mandlekar2022robomimic,kim2024openvla,pertsch2025fast}. Generative imitation instead fits an action distribution: diffusion learns iterative denoising, while flow matching learns a velocity field that transforms noise toward action samples~\citep{chi2023diffusionpolicy,black2024pi0}. Policy distillation instead transfers a teacher's behavior to a student~\citep{wang2025onedp}. And adversarial distribution matching provides another route to imitation~\citep{ho2016gail}. The learning objective can also be the improvement target~\citep{yu2018daml}, which provides a methodological basis. But evidence of physically grounded, persistent adaptation of IL mechanisms remains limited.

\textbf{Self-supervised learning} (SSL) derives training targets from observations and trajectories by delving into their internal structure and relationships. Reconstruction and latent-action objectives learn visual or motion representations without expert action labels, although mapping them to robot control may require supervised adaptation~\citep{radosavovic2022maskedvisual,ye2024lapa}. Predictive learning extends this approach to environmental dynamics~\citep{mendonca2023swim}. In a physical improvement loop,  robot trajectories provide self-supervised targets for updating a predictive model, which then generates imagined experience for subsequent policy training~\citep{wu2023daydreamer,guo2026vlaw}. Self-supervised updates can also adapt policy memory during execution: RoboTTT updates fast weights to incorporate incoming observations into the context used for subsequent actions~\citep{jiang2026robottt}. At the mechanism level, AutoSSL searches combinations and weights of self-supervised objectives, but does not incorporate a physical robot learning loop~\citep{jin2022autossl}.

\textbf{Reinforcement learning} (RL) enables robots to improve through repeated interaction: execution outcomes guide policy or value updates, which influence subsequent actions and learning~\citep{hirose2025selfi,stachowicz2025liren,mendonca2025continually,wang2026lwd}. Offline data and demonstrations provide a starting point, while online interaction and human corrections support later updates~\citep{nakamoto2023calql,prsi_conrft,amin2025pi06}. The learned component may be the base policy or a steering policy around a frozen behavioral prior~\citep{ren2024dppo,wagenmaker2025dsrl}. In the improvement loop, the reward signal can be adjusted through online reward-model updates or iterative reward-code revision~\citep{zhang2026evohil,prsi_roska}. Mechanism-level improvement further targets how learning updates or rewards are generated: DiscoRL learns the rules for updating policies and predictions, while DERL uses policy validation performance to train the meta-optimizer that generates future rewards~\citep{oh2025discorl,cheng2025derl}.

\subsection{Learning Models}

The learning methods above apply to diverse models used in robot policies and their training. This section focuses on VLA models, world models, and reward models, which have a strong influence on continual improvement by shaping execution and subsequent learning. VLA updates change actions and collected experience; world-model updates change predictions used for planning or imagined training; reward-model updates change the rewards or value estimates guiding subsequent policy learning.

\textbf{Vision-language-action} (VLA) models generate robot commands from visual observations, language instructions, and optional sensory inputs~\citep{kawaharazuka2025vlasurvey}. Continual adaptation incorporates demonstrations, corrective interventions, and deployment trajectories into IL or RL updates~\citep{prsi_conrft,amin2025pi06}. One route updates internal components, including action heads, adapters, backbone layers, or the full model~\citep{kim2024openvla}. Action heads map multimodal representations to discrete tokens or action chunks~\citep{zitkovich2023rt2,black2024pi0}; continuous generators include diffusion heads and diffusion Transformers~\citep{ghosh2024octo,liu2025rdt1b}. Updating these components changes action generation, while backbone adaptation can also change visual and instruction grounding. Knowledge Insulation controls this coupling by adapting the backbone through discrete-action supervision while blocking gradients from the continuous-action expert~\citep{driess2025knowledgeinsulation}. Within hierarchical VLAs, internal adaptation may target task interpretation, action generation, or both~\citep{belkhale2024rth,shi2025hirobot,bjorck2025gr00tn1}. A second route keeps a frozen VLA and learns a lightweight controller from physical experience. FIND learns residual corrections using outcome assessment and human assistance for scene recovery~\citep{fang2026find}. RLT instead trains a lightweight actor critic on an RL token from the frozen VLA, refining action chunks while anchoring the actor to VLA recommendations~\citep{xu2026rlt}. Here, improved behavior is carried by the controller instead of the VLA weights.

\textbf{World models} learn action-conditioned predictions of future observations, latent states, or object relations~\citep{hou2026worldmodelsurvey}. Robot interaction can adapt predictive priors learned from human video to the deployment environment~\citep{mendonca2023swim}. Their role in continuing improvement depends on whether these predictions remain fixed or are updated with new experience. A fixed world model can support repeated policy optimization through imagined rollouts~\citep{yang2026rise}. Alternatively, new policy rollouts update the world model, which then provides predictions for subsequent policy training~\citep{liu2026worldvlaloop}. The predictor then becomes an evolving source of training experience. Updated predictions can change the actions inferred through inverse dynamics~\citep{du2023unipi}, the ranking of plans or candidate behaviors~\citep{hansen2024tdmpc2,prsi_worldeval}, and the synthetic experience used for policy training~\citep{yang2024unisim}. These uses require accurate predictions of action outcomes, beyond visually realistic observations~\citep{guo2025ctrlworld}. In WAMs, prediction and action generation are integrated within a single model, so world-model adaptation can also affect control~\citep{bi2026motus2,qian2026wamrl}.

\textbf{Reward models} learn to score task progress, success, or relative behavior quality from videos, outcome labels, and trajectory comparisons~\citep{ma2023liv,lee2026roboreward,liang2026robometer}. They support continuing improvement through two distinct paths. A fixed evaluator supplies rewards for successive policy updates, with dense progress estimates complementing sparse task outcomes~\citep{chen2026sarm2,fang2026densereward,prsi_dopamine}. Alternatively, the evaluator itself learns from new feedback. Reward-model updates can alternate with policy learning, requiring stored experience to be rescored as the learned rewards change~\citep{lee2021pebble}. In physical deployment, reward models can be adapted by human confirmed success used for subsequent policy training~\citep{zhang2026evohil}. Adaptation to new preferences can be confined to low-rank modules, reducing interference with earlier reward knowledge while keeping the base reward-model weights fixed~\citep{marta2025flora}. Critic updates provide another form of evaluator adaptation: new successful and failed rollouts refine expected-return estimates used to select actions. This can improve value-guided action selection while the underlying action-generating policy remains frozen~\citep{giridhar2026imitationselfimprovingrobotpolicies}.

\subsection{Knowledge Transfer and Training Procedure Improvement}
\label{sec:training-continual}
\label{sec:training-procedures}

Across learning rounds, accumulated knowledge can support new tasks, while feedback from training and execution can inform how subsequent models are trained. We review these routes through cross-task transfer from learned capabilities, followed by feedback-driven improvements to data preparation and training methods.

\textbf{Cross task transfer} uses knowledge acquired on source tasks to support target task learning, while preserving the previously acquired capabilities. Policy initialization starts target task training from a multi-task generalist or a source task policy selected using predicted zero shot success~\citep{bousmalis2024robocat,guo2025srsa}. Skill and module reuse instead incorporates learned skill policies or temporal action representations into the target policy, with learning focused on their selection, composition, or adaptation~\citep{zhang2025extract,zheng2024prise}. To adapt these reusable components to a target task, learning can focus on skill selection, expert routing~\citep{wan2024lotus,wang2024sparsediffusion}, or controller composition~\citep{feng2026divmorph}. Throughout updates, replay, modular adaptation, and parameter merging help preserve earlier knowledge while incorporating target-task experience~\citep{he2026lifelongvla,prsi_merging,liu2026pretrainedvla,zhu2026continualvla}.

\textbf{Data preparation improvement} uses learning and execution outcomes to revise what experience is prepared for subsequent training. Successful demonstrations do not guarantee robots can reliably learn the demonstrated behaviors. Feedback from policy learning and execution can help identify suitable demonstrations for subsequent training~\citep{chen2025demoscore, agia2025cupid,dass2025datamil}. Learning feedback can also guide how data from different sources are mixed for subsequent policy training~\citep{hejna2024remix} and informs future experience generation: real-world policy returns guide simulation distributions, while fresh rollouts update predictors that direct collection toward current failure regions~\citep{muratore2021bayrn,he2026criticality}. These approaches connect data decisions to learner performance. Extending this feedback to autonomous strategy revision, RSIBench-Data evaluates agents that use training and evaluation results to revise data sources, coverage, trajectory representations, and curation strategies in digital tasks~\citep{meng2026rsibenchdata}.

\textbf{Training method improvement} changes training methods for later learning, including revisions to reward code, the training distribution, the update rule, or the organization of training experiments~\citep{prsi_eureka,chebotar2019simopt,oh2025discorl,chen2026evotrainercoevolvingllmpolicies}. Robot rollout results can guide revisions to training code and recipes~\citep{xiao2026enpire}, while records of earlier tuning attempts can guide later training decisions~\citep{li2026harborharnessframeworkagentic}. These changes enter the self-improvement loop when the system retains and uses them to train subsequent models. Physical RSI additionally requires training-mechanism revisions to be \emph{physically grounded}: feedback from the robot's interaction with its environment must inform how the training method changes. Retaining and reusing such a revision can support L2 on the same task or L4 across tasks. L5 further requires showing that the revised mechanism yields better subsequent improvements than its predecessor under matched starting capabilities and resources. Section~\ref{sec:physical-rsi-evaluation} describes how to test this gain.

\begingroup
\raggedbottom

\section{Safety and Control of Physical RSI}
\label{sec:safety_control}
\label{sec:safety}

Physical RSI must preserve safety as both the task-execution system $M=(\Theta,H)$
and the improvement procedure $\Phi$ evolve. System updates can change which
states the system visits and alter its hazard estimates and action interfaces.
Environmental drift can change
the dynamics and available control responses. Revisions to $\Phi$ can further
change how trials are chosen and updates are accepted, allowing errors in safety
assessment to influence later generations.

The reviewed literature mainly provides component results, with limited evidence
on safety across inherited revisions of $\Phi$. Following
Section~\ref{sec:architecture}, safety constrains the execution, capability-level
update, and meta-level update loops. Safety requirements and the independent
evaluator $E$ with its scoring criterion $Q$ remain protected from $\Phi$ even
when safeguard implementations change. Under these constraints, control includes
restricting actions and trials, specifying permitted operating conditions, and
retaining the ability to interrupt execution or improvement. The question
throughout this section is which safety evidence remains valid after a change
and what must be checked before further physical
operation (Figure~\ref{fig:safety_control_overview}).

\begin{figure*}[t]
    \centering
    \includegraphics[width=\textwidth]{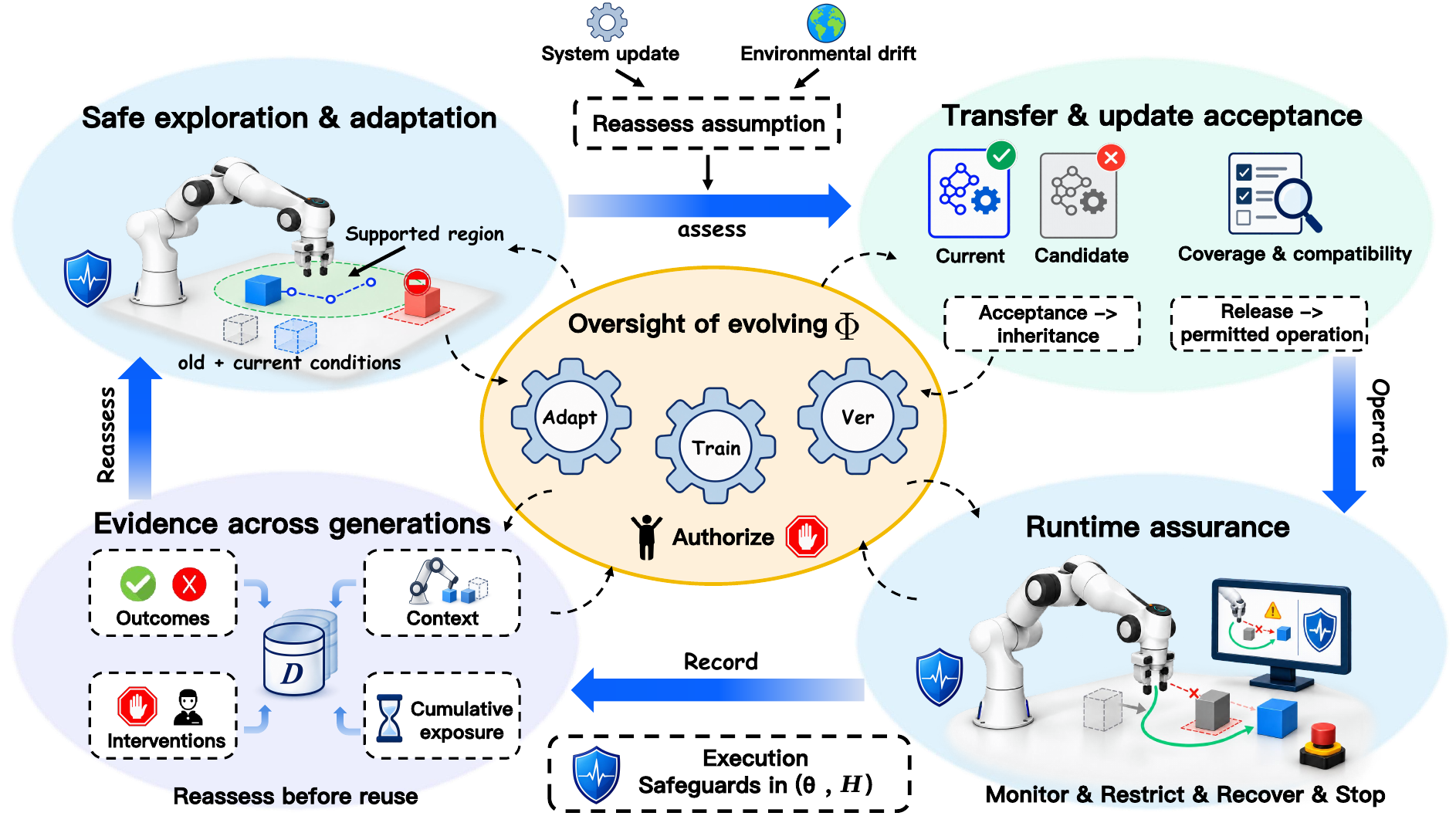}
    \caption{Safety and control in Physical RSI. System updates and environmental drift require reassessment of exploration, transfer, and runtime safeguards. Records in $D$ inform later revisions of $\Phi$. Acceptance determines inheritance, while release specifies permitted physical operation. Oversight of the evolving procedure spans the cycle.}
    \label{fig:safety_control_overview}
\end{figure*}

\subsection{Safe Exploration and Continual Adaptation}
\label{sec:safety_exploration}

\paragraph{Safety during learning.}
A policy that is safe after training may still cause harm while collecting experience. Safe reinforcement learning (RL) therefore distinguishes the constrained quantity, the stage at which it is constrained, and the supporting evidence \citep{gu2024safereview,gu2023safe}. In a constrained Markov decision process, rewards measure task performance and costs represent safety violations. Constrained Policy Optimization approximately maintains expected-cost limits during policy updates \citep{achiam2017cpo}. SafeVLA applies constrained learning to unsafe behaviors of vision-language-action models \citep{zhang2025safevla}. Other methods address conflicts between reward and safety gradients
\citep{gu2024pcrpo}, balance multiple objectives under safety constraints
\citep{gu2025safe-balance}, or incorporate risk measures into safety bounds
\citep{gu2025roscom}. Expected-cost limits can still permit individual violations, and their meaning depends on whether the cost captures the physical hazard.

\paragraph{Retaining safety while adapting.}
Safe Continual Domain Adaptation (SCDA) combines safe RL with a penalty on changes to important policy parameters after pretraining across varied simulated conditions \citep{josifovski2025scda}. Its robot experiments show adaptation with reduced forgetting and controlled average costs under the tested conditions. The penalty constrains the policy, while learned estimates of reward and safety cost continue adapting. Fitting these estimators to prescribed targets does not by itself establish a meta-level update, even when their parameters persist across rounds (Section~\ref{sec:levels}).

Retaining task performance does not necessarily retain safety. In simulated non-stationary benchmarks, \citet{coursey2026safecontinual} find that reducing forgetting does not reliably preserve safety performance, while stronger cost penalties can reduce task success. Post-update evaluation should therefore cover both previous and current operating conditions. Historical safety labels should also be reassessed against current dynamics and sensing conditions.

\paragraph{Exploration within a supported region.}
Gaussian-process controller optimization restricts quadrotor trials to settings predicted to meet a safety threshold \citep{berkenkamp2016quadrotor}. Sequential safe exploration additionally checks reachability and return paths \citep{turchetta2016safe}. Stability certificates support joint improvement of a controller and its estimated safe region under regularity and uncertainty assumptions \citep{berkenkamp2017stability}.

These results rely on assumptions that may no longer hold after an update to $M$ or $\Phi$, or a change in the environment. Before further exploration, the affected cost estimates, uncertainty bounds, and return paths must therefore be checked against current physical conditions. The supported region may need to shrink under drift. If no available trial supports safe continuation under these assumptions, physical exploration must pause.

\subsection{Safe Transfer and Update Acceptance}
\label{sec:safety:acceptance}

\paragraph{What transfers with a policy.}
Transfer of capability does not establish that safety evidence transfers with it. Sim-to-Lab-to-Real combines a learned backup policy with restrictions on unsafe actions and statistical bounds on expected performance and safety \citep{hsu2023simlabreal}. Its generalization argument assumes that laboratory and deployment environments follow the same distribution. Robust safe transfer instead uses a guide policy whose tolerance to uncertainty covers the target environment \citep{zubia2025robusttransfer}. Its expected-cost guarantee requires compatible action spaces and a state mapping that preserves safety conditions.

These assumptions concern the system and environment together. A new encoder can invalidate the state mapping, while an interface change can invalidate the guide's actions. Neither change necessarily alters the required skill sequence $\kappa$, so safety reassessment also applies within a task as defined in Section~\ref{sec:levels}.

\paragraph{Evidence for accepting an update.}
Under Sections~\ref{sec:architecture} and~\ref{sec:evaluation-verification}, $\mathrm{Ver}$ internally judges changes to $M$ and to $\mathrm{Adapt}$ or $\mathrm{Train}$. The independent evaluator $E$, using the protected criterion $Q$, measures gains in the updated task-execution system and assesses revisions to $\mathrm{Ver}$ by evaluating the systems retained under each verifier. Acceptance determines inheritance. Establishing improvement requires independent outcome evaluation (Section~\ref{sec:evaluation-outcomes}). Release is the separate authorization for physical operation under specified conditions and safeguards, including restricted trials used to collect evidence.

High Confidence Policy Improvement bounds the probability of accepting a policy below a return threshold \citep{thomas2015hcpi}. The Seldonian framework tests specified behavioral constraints and can return no solution when evidence is insufficient \citep{thomas2019seldonian}. Constrained offline learning and risk-sensitive objectives account for costs and severe outcomes beyond average return \citep{lee2022coptidice,chow2015risk}. Logged data can support policy evaluation and limit policy changes where evidence is sparse \citep{jiang2016doubly,laroche2019spibb}. The validity of these guarantees and estimates depends on data coverage and estimation assumptions, which must be reconsidered under non-stationarity \citep{chandak2020nonstationary}. A gain over an unsafe baseline alone is insufficient.

\paragraph{Releasing a compatible system.}
The policy, observation processing, action interface, monitor, and backup controller should be assessed together. Harness changes require checks of skill preconditions, invocation timing, and handover states (Sections~\ref{sec:harness-skills} and~\ref{sec:harness-verification}). Changes to execution models require checks of compatibility with the retained harness (Section~\ref{sec:training-method}). Hardware checks include sensing errors, actuator limits, and control timing. Safety filters can account for bounded sensing errors, as demonstrated by measurement-robust barrier functions in camera-based Segway control \citep{cosner2021measurement}.

Definitions~\ref{def:psi} and~\ref{def:prsi} allow physical grounding through interaction in a physics simulator or the real world. This grounding condition is distinct from evidence that a particular hardware deployment is safe. Learned world-model predictions can help screen candidates but cannot supply physical grounding on their own. Simulation evidence alone does not establish hardware safety, which requires assessment of the deployed system under its intended physical conditions.

After a system or procedure update, release requires evidence covering the intended conditions and the complete component combination. Improving the score on $E$ does not supply this evidence by itself. The assessed conditions must remain explicit so that runtime monitoring can detect when they no longer hold.

\subsection{Runtime Assurance under Environmental Drift}
\label{sec:safety_recovery}

\paragraph{Preserving a feasible response.}
Runtime assurance restricts actions or invokes a backup controller when continued execution threatens safety. Safeguards used during task execution belong to $M$: learned model weights are part of $\Theta$, while filter, switching, and controller code belong to $H$. Components used only within the improvement procedure belong to $\Phi$, while the independent evaluator $E$ remains outside it (Section~\ref{sec:definition}).

Control barrier functions preserve a safe region under model and control-feasibility assumptions \citep{ames2017cbf}. Action filters can incorporate learned dynamics and uncertainty estimates \citep{dalal2018continuous,cheng2019barrier}. Reachability methods account for modeled disturbances \citep{fisac2019framework}, while predictive filters check future constraints and safe continuation \citep{wabersich2021filter,tearle2021racing}. Shielding restricts actions to satisfy logical specifications, including checks that those specifications can be met with the available controls \citep{alshiekh2018shielding,kim2025shields}. Removing a filter after training removes its direct runtime enforcement \citep{yang2026cbfrl}. Logs must distinguish proposed and executed actions, since frequent corrections can conceal an unsafe policy.

\paragraph{Detecting failure and invalid evidence.}
Failure prediction supplies warning signals but does not itself enforce physical constraints. Visual predictors can bound false alarms and missed failures for defined outcomes \citep{farid2022failure}. Uncertainty in language-model judgments can support deferring uncertain decisions \citep{zheng2024evaluating}. Sentinel checks execution consistency and progress \citep{agia2025sentinel}. SAFE detects task failures from VLA features. Under the exchangeability
assumption, its conformal calibration bounds false alarms on successful
trajectories, but does not guarantee detection of failures or the safety
of intervention \citep{gu2025safe}.

DoReMi and Code-as-Monitor connect execution deviations to replanning or geometric checks \citep{guo2024doremi,zhou2025codeasmonitor}. Other detectors distinguish unusual observations from likely failures \citep{romer2025fiper,rolland2026fidel}. Policy and encoder updates can change both the visited states and the features used by these detectors. Recalibration needs evidence from the updated system. Raising alarm thresholds merely to reduce interventions can conceal declining safety.

\paragraph{Recovery after change.}
SOTER and Neural Simplex combine switching rules with backup control \citep{desai2019soter,phan2020neural}. Recovery RL learns recovery behavior from estimates of future violations \citep{thananjeyan2021recovery}. Such methods require a recoverable state and an effective controller. Restoring an earlier policy cannot reverse a mechanical fault or changed friction. Learned resetting and return behaviors support autonomous practice but also require physical trials and can fail \citep{eysenbach2018leave,sharma2022medal,sharma2023selfimproving}. In shared workspaces, coordinated control must remain feasible when other robots change their behavior \citep{wang2017multirobot}.

Changes to a policy, monitor, or backup controller require joint checks of detection errors, response delays, and recovery feasibility. Revisions to $\Phi$ must carry these checks into subsequent updates. Monitoring must track the physical assumptions as well as task failures, and restrictions must act before recovery becomes infeasible. Abrupt changes outside the assumed disturbance bounds remain outside the guarantee.

\subsection{Oversight of an Evolving Improvement Procedure}
\label{sec:safety_human}

\paragraph{Constraining how successors are produced.}
Changes to $\mathrm{Adapt}$, $\mathrm{Train}$, or $\mathrm{Ver}$ affect the trials used to develop later systems and the decisions that retain them. Omitting intervention records from $D$ can make assisted task success appear to show that a policy is safe on its own. Procedure assessment must therefore include the physical trials and subsequent systems it produces, rather than only the final policy's performance.

Formal guarantees cover some structured updates. \citet{fulton2019offmodel} revise models, control logic, and associated proofs, then eliminate inconsistent models during operation. Their safety result requires an accurate model to remain in the candidate set. This supports the specified model and controller updates, not arbitrary revisions of $\Phi$.

\paragraph{Preserving requirements and evaluation.}
A revised cost estimator must still measure the hazards specified by the protected requirements. If one learned model both proposes actions and assesses their safety, shared prediction errors can affect both decisions. A verifier can also drift toward accepting updates that pass its local checks while increasing physical risk. Changes to $\mathrm{Ver}$ therefore require assessment through protected $E,Q$ and evidence from the systems retained under the competing verifier versions (Section~\ref{sec:evaluation-verification}). The safety criteria for accepting a revision of $\mathrm{Ver}$ should be specified before testing and protected from modification by the candidate verifier. Passing these safety checks does not by itself establish downstream recursive gain, which requires the controlled comparisons of Section~\ref{sec:evaluation-outcomes}.

Instructions embedded in external content can redirect an agent \citep{greshake2023indirect}. When $\Phi$ reads logs or edits code, this content must not acquire permission to change protected requirements or disable stopping mechanisms.

\paragraph{Human control and autonomous inheritance.}
Human demonstrations and corrections may enter $D$ under the system description in Section~\ref{sec:definition}. Their use must nevertheless be distinguished from the autonomy required for a level claim. Section~\ref{sec:levels} assigns workflows whose retained changes depend on human intervention or labeling during improvement to L0. Direct human edits to $\Theta$, $H$, or $\Phi$ also violate the self-inheritance condition, and human selection of retained updates replaces autonomous retention. Authorization for physical operation is a separate safety decision. Its role, and any interventions that contribute to learning or update selection, must be recorded explicitly (Section~\ref{sec:evaluation-outcomes}).

Human interventions provide experience for policy learning \citep{kelly2019hgdagger,liu2023sirius,luo2025hilserl,chen2025conrft}, although uncertain or suboptimal guidance needs additional care \citep{zhao2025silri}. Emergency stops should remain distinguishable from demonstrations of successful behavior. Methods for allocating human attention and requesting clarification can reduce supervision demands \citep{hoque2022thrifty,hoque2023fleet,ren2023knowno}, but do not ensure help arrives before danger. AutoRT and SOP provide supervision and intervention infrastructure \citep{ahn2024autort,pan2026sop}. Work on interruptible learning and the Off-Switch Game addresses incentives around interruption \citep{orseau2016interruptible,hadfield2017offswitch}. Physical intervention also depends on sensing, communication, actuation, and response time.

Before inheriting a revision of $\Phi$, assessment must check how it changes trial selection, interpretation of interventions, and update acceptance. Protected requirements and effective stopping mechanisms must persist through the change. Replacing a safeguard additionally requires compatible controllers and a recoverable physical state at handover. Human involvement in operational safety and in inheritance decisions must be recorded separately.

\subsection{Safety Evidence across Generations}
\label{sec:safety_implications}

Each generation needs records linking its system and procedure versions to the physical conditions, interventions, and outcomes behind its safety assessment. These records allow later changes to be checked against the assumptions on which earlier decisions relied. Table~\ref{tab:safety_evidence} summarizes the evidence available for different changes and the checks needed before reuse.

\begin{table*}[!t]
    \centering
    \caption{Safety evidence and reassessment under system and environment changes.}
    \label{tab:safety_evidence}
    \small
    \setlength{\tabcolsep}{6pt}
    \renewcommand{\arraystretch}{1.05}
    \begin{tabular}{@{}
        >{\raggedright\arraybackslash}p{0.30\textwidth}
        >{\raggedright\arraybackslash}p{0.35\textwidth}
        >{\raggedright\arraybackslash}p{\dimexpr0.35\textwidth-4\tabcolsep\relax}
        @{}}
        \toprule
        \textbf{Change and examples} & \textbf{Evidence and limits} & \textbf{Checks before reuse} \\
        \midrule
        \textbf{Policy adaptation}\newline
        SCDA \citep{josifovski2025scda} &
        Empirical cost control and reduced forgetting in tested conditions. &
        Costs in old and current conditions, including adaptation trials. \\
        \addlinespace[4pt]
        \textbf{Domain transfer}\newline
        \citep{hsu2023simlabreal,zubia2025robusttransfer} &
        Statistical or expected-cost bounds under transfer assumptions. &
        Target coverage, state mappings, and action compatibility. \\
        \addlinespace[4pt]
        \textbf{Dynamics or interfaces}\newline
        Barrier functions \citep{ames2017cbf} &
        Safe-region preservation under valid models and feasible control. &
        Sensing, actuator limits, control timing, and handover state. \\
        \addlinespace[4pt]
        \textbf{Monitors and recovery}\newline
        \citep{gu2025safe,thananjeyan2021recovery} &
        False-alarm calibration or learned recovery estimates. Intervention can still fail. &
        Feature validity, missed hazards, response delays, and recovery feasibility. \\
        \addlinespace[4pt]
        \textbf{Structured updates}\newline
        \citep{fulton2019offmodel} &
        Proof-preserving model and controller updates within a specified family. &
        Protected rules and the trials and systems produced by the revision. \\
        \addlinespace[4pt]
        \textbf{Repeated selection}\newline
        \citep{dwork2015generalization} &
        Statistical validity depends on controlling adaptive reuse of feedback. &
        Statistical error control across rounds, with physical exposure accounted for separately. \\
        \bottomrule
    \end{tabular}
    \par\smallskip
    {\footnotesize Checks are our synthesis. The cited results do not establish safety for arbitrary revisions of $\Phi$.\par}
\end{table*}

\paragraph{Risk across rounds.}
A bound for one round does not establish a lifetime guarantee. If $V_k$ denotes a physical violation in round $k$ and valid bounds $\Pr(V_k)\leq\delta_k$ hold for the adaptive process, then
\[
    \Pr\!\left(\bigcup_{k=1}^{K}V_k\right)
    \leq \sum_{k=1}^{K}\delta_k .
\]
This union bound needs no independence, but its per-round bounds must remain valid for the adaptive process. Estimates from obsolete models cannot supply them. Physical risk accumulates during rejected trials, recovery, and resetting, while hardware wear can alter the assumptions used in later assessments. Risk incurred while developing a revision of $\Phi$ should be recorded separately from risk during its subsequent use, alongside the resource accounting discussed in Section~\ref{sec:evaluation-outcomes}.

False acceptance of an update is distinct from physical harm. Repeated selection needs statistical error control, and adaptive reuse of test feedback requires fresh evidence or a correction for the resulting dependence \citep{dwork2015generalization}. In the architecture of Section~\ref{sec:architecture}, judgments from $\mathrm{Ver}$ or $E$ are recorded in $D$. If external judgments influence later proposals or candidate acceptance, the associated evidence cannot also serve as untouched final evaluation. Records must distinguish feedback used during improvement from evidence reserved for independent assessment (Section~\ref{sec:evaluation-outcomes}). Protecting evaluation tasks and $Q$ from direct access or modification does not by itself remove adaptive selection effects.

\paragraph{Evaluation after change.}
Safety evaluation should cover adaptation, rejected trials, resets, and recovery, reporting violation frequency and severity, near misses, missed failures, intervention delays, and recovery success. Tests immediately after updates reveal failures hidden by final-policy averages. Revisiting previous conditions tests safety retention. Comparing fixed and updated systems under both stable and drifting conditions helps distinguish environmental effects from update effects.

Safe operation and recursive gain require distinct evidence. For L5, each assessed revision of $\Phi$ must be autonomously inherited and produce better subsequent improvements across tasks than its predecessor. The matched comparison and independent evaluation requirements are specified in Definition~\ref{def:prsi} and Section~\ref{sec:evaluation-outcomes}.

For safety, these comparisons must also match safety rules, stopping and reset conditions, and the arrangements for human oversight and authorization for physical operation. Actual interventions and deviations from these conditions must be recorded, together with risk incurred during adaptation and recovery. Repeated trials under matched operating conditions help account for physical variability.

Safety evidence can be reused only while the relevant assumptions and statistical bounds remain valid. Each inherited change must therefore be linked to checks of the affected safeguards and to the physical risk incurred in producing and using it. These checks support assessment under stated conditions, not a guarantee for arbitrary environmental drift or unrestricted Physical RSI.

\par
\endgroup

\begin{figure*}[t]
    \centering
    \includegraphics[width=1\linewidth]{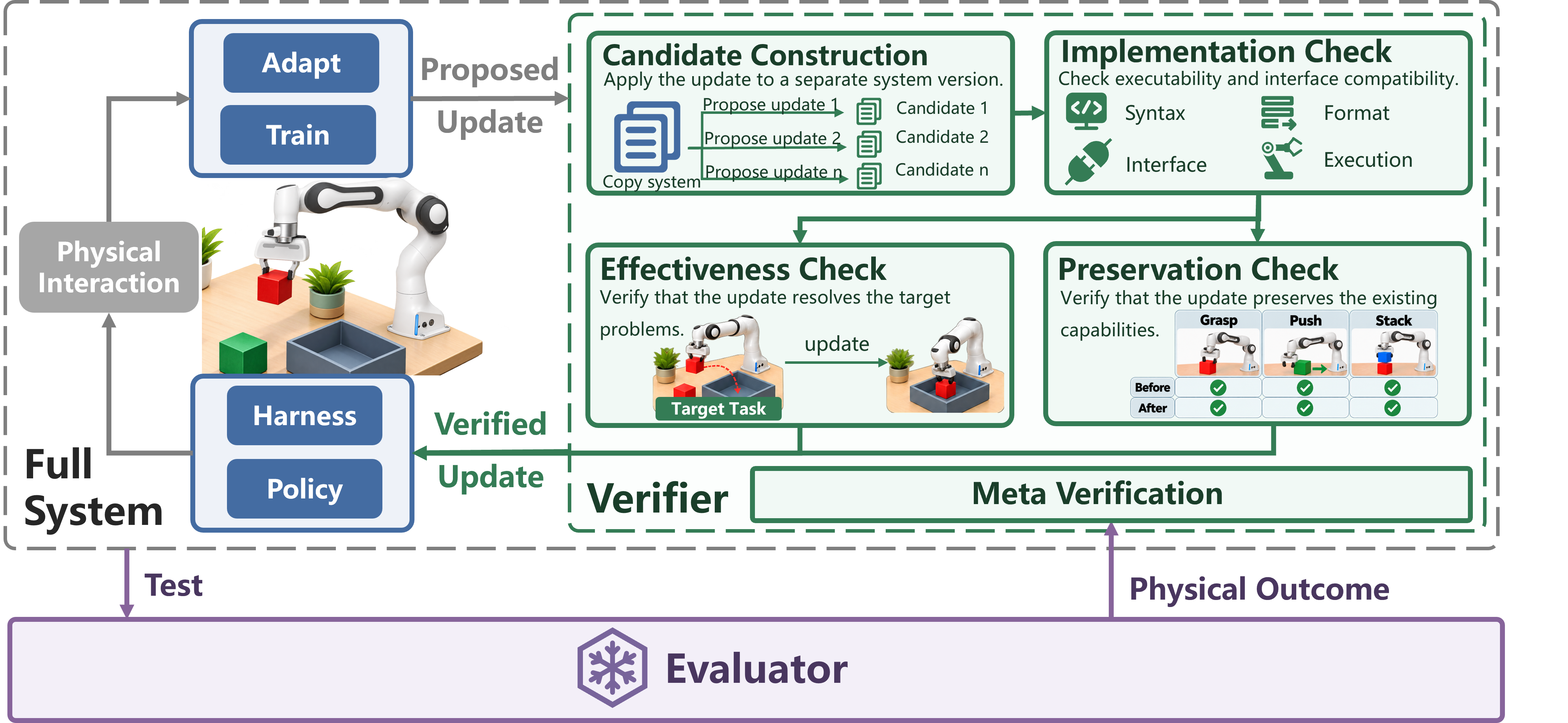}
    \caption{Verification and evaluation in physical RSI. Proposed updates are applied to separate candidate system versions after harness adaptation and model training. Verification checks implementation validity, repair effectiveness, and capability preservation before updates are adopted. An independent evaluator assesses the updated system under external test conditions and scoring criteria, providing physical outcome evidence for meta-verification when the verifier itself is revised.}
    \label{fig:ver_eval}
\end{figure*}
\section{Verification and Evaluation of Physical RSI}
\label{sec:physical-rsi-evaluation}
\label{sec:evaluation}

Evaluating Physical RSI requires assessing both whether candidate updates pass verification and whether their subsequent use improves task performance or the improvement procedure. Following Section~\ref{sec:definition}, verification within the improvement procedure $\Phi$ examines whether proposed updates should be accepted and inherited, while independent evaluation uses an external $E$ and criterion $Q$ to assess the updated task-execution system $M=(\Theta,H)$. Passing internal verification does not by itself establish subsequent improvement, which requires independent evaluation. As illustrated in Figure~\ref{fig:ver_eval}, this section reviews verification practices for candidate updates (Section~\ref{sec:evaluation-verification}), evaluation of improvement outcomes (Section~\ref{sec:evaluation-outcomes}), and related benchmarks (Section~\ref{sec:physical-rsi-benchmarks}).

\subsection{Verification of Candidate Updates}
\label{sec:evaluation-verification}
\label{sec:evaluation-process}

Verification checks whether a candidate update is ready for adoption by examining its implementation, its effect on the target problem, and its impact on existing capabilities. Following Section~\ref{sec:definition}, $\mathrm{Ver}$ is part of the improvement procedure $\Phi=(\mathrm{Adapt},\mathrm{Train},\mathrm{Ver})$ and informs which updates to $M=(\Theta,H)$ or to the improvement procedure are accepted and inherited. These internal checks use development evidence, instead of replacing the protected external evaluation $E$ under criterion $Q$.

\textbf{Implementation Validity.} Implementation checks examine whether a candidate can run correctly within the system, including syntax, interfaces, formats, and execution outputs. Checks at stage boundaries or before task execution identify implementation errors before more costly task tests, as illustrated by HARBOR and HarnessPAI~\citep{li2026harborharnessframeworkagentic,agentdraft_hsa_harnesspai2026}. Passing these checks establishes that the candidate can be tested; its task effect still requires execution evidence.

\textbf{Repair Effectiveness.} Repair checks assess whether a modification resolves the targeted problem. Re-execution on failed cases and comparison with preceding versions on matched task instances provide evidence through outcomes and execution traces~\citep{lu2026aspire,prsi_rho}. These tests connect the proposed correction to its measured effect within tested conditions. Applying an update to other tasks requires evidence of its effectiveness beyond the original case: RoboFoundry checks general updates by generating repair proposals from other pre-repair records and assessing them against previously successful repair updates and their execution outcomes~\citep{agentdraft_liang2026robofoundry}.

\textbf{Capability Preservation.} Preservation checks revisit successful tasks or conditions to detect losses caused by a shared modification. Acceptance criteria may require previously successful tasks to remain successful or restrict performance losses on individual tasks. RPG and HarnessPAI retest the updated systems before accepting changes to check that previously successful tasks remain successful~\citep{wang2026rpg,agentdraft_hsa_harnesspai2026}. Preserving existing capabilities also requires checking whether previously learned knowledge remains valid after an update: KnowBody suspends rules that depend on revised body estimates until they are revalidated \citep{agentdraft_lou2026knowbody}.

\textbf{Meta-verification.} When $\mathrm{Ver}$ itself is revised, its acceptance rules become part of the object under examination. This evaluation matters because verification mechanisms used to assess candidate updates can themselves make incorrect judgments,  as documented by FailBench and \emph{No Free Checker}~\citep{navasardyan2026failbench,prsi_nofree}. VERSE provides a digital example of developing verification-related tools around fixed checking primitives \citep{agentdraft_hsa_verse2026}. Such revisions cannot be judged solely by their own acceptance standards: accepting more updates does not establish that a revised verifier is better. Independent evaluation should therefore compare the subsequent performance of systems produced under the original and revised verification procedures, using a common external evaluation set $E$ and scoring criterion $Q$.

\subsection{Evaluation of Improvement Outcomes}
\label{sec:evaluation-outcomes}
\label{sec:evaluation-metrics}

Whereas verification determines which candidate updates are accepted, independent evaluation uses the protected evaluator $E$ and scoring criterion $Q$ (Section~2.1) to assess the effects of changes retained in later task-execution systems $M=(\Theta,H)$. Revisions to the improvement procedure $\Phi$ are assessed indirectly through the subsequent systems they produce. We review how existing studies assess performance across improvement rounds, identify the contributions and transferability of updates, and report resource costs and human involvement.

\textbf{Performance and Retention.} Existing evaluations examine both the gains from a single learning update and how performance evolves across successive improvement rounds. Before--after comparisons measure changes in task success~\citep{soar,yu2026f4r}. Multi-round evaluations extend this comparison to successive policy checkpoints or updated agent systems, tracking whether gains accumulate, persist, or temporarily decline~\citep{lang2026embodirsi,wang2026rpg}. To assess these changes independently, updated systems are evaluated under initial conditions that were not used for debugging or selecting candidate updates~\citep{lu2026aspire, wang2026rpg}.

\textbf{Attribution and Transfer.} Controlled comparisons identify which changes contribute to improvement, while transfer tests examine their usefulness beyond single tasks. EmbodiRSI compares adaptive and static collection with matched trajectory counts, initial data, policy architectures, and training recipes~\citep{lang2026embodirsi}. RPG ablates diagnostic inputs and separately compares skill-library versions with the runtime model, prompt, perception, and execution budgets fixed~\citep{wang2026rpg}. ASPIRE evaluates library snapshots accumulated from different numbers of source tasks, including an empty-library baseline on unseen tasks; it also tests simulation-discovered skills on a different physical embodiment~\citep{lu2026aspire}. These evaluations connect gains to specific changes and assess whether their benefits can be reused beyond the original tasks and embodiments.

\textbf{Efficiency and Autonomy.} Several studies report the resources behind improvement and distinguish autonomous execution from assistance during development. ENPIRE measures robot and GPU utilization, token consumption, and time and tokens needed to reach success, exposing trade-offs between faster improvement and resource efficiency~\citep{xiao2026enpire}. KnowBody reports planner rounds alongside task completion and tracks decision effort during continual interaction~\citep{agentdraft_lou2026knowbody}. F4R compares methods under matched data-preparation time budgets and itemizes reconstruction, collection, and training costs~\citep{yu2026f4r}. Autonomy is assessed through explicit execution conditions: EmbodiRSI disables human correction and agent takeover during policy evaluation, while RPG distinguishes autonomous practice updates from human setup and hardware adaptation, then evaluates frozen physical systems without human assistance~\citep{lang2026embodirsi,wang2026rpg}. These reports relate improvements to resource use and human involvement.

% Taken together, these practices support claims about independently
% measured capability gains, their sources and scope, and the resources
% required to obtain them. They do not, however, establish the recursive
% gain of Definition~\ref{def:prsi}. That stronger claim requires an
% autonomously inherited revision $\Phi_k$ to produce better subsequent
% improvements than $\Phi_{k-1}$ from the same starting generation and
% experience, under the same budget and protected $E,Q$, with gains
% measured across tasks. The reviewed robot studies provide more direct
% evidence for capability improvement and component-level attribution
% than for this controlled, procedure-level comparison, which remains a
% central evaluation gap for Physical RSI.

\subsection{Benchmarks and Evaluation Resources}
\label{sec:physical-rsi-benchmarks}

Table~\ref{tab:rsi-benchmark-levels} organizes existing resources into embodied and RSI-related benchmarks according to their primary evaluation focus. The resources include task suites, benchmark protocols, and supporting evaluation frameworks; a resource may support more than one focus.

\textbf{Embodied benchmarks.} Task-oriented benchmarks assess task learning, performance, and generalization~\citep{liu2023libero}, as well as robustness and physical-world performance~\citep{zhou2025liberopro,pmlr-v305-atreya25a}. Alongside task-level evaluations, supporting resources examine the reliability of success and reward judgments~\citep{navasardyan2026failbench,lee2026roboreward} and world-model-based policy assessment~\citep{tseng2026sc3eval}. Evaluation also extends to the development process, assessing robot programming and policy-training workflows~\citep{pmlr-v306-fu26u,lei2026embocoachbench}.

\textbf{RSI-related benchmarks (primarily digital).} Recent resources examine agent-driven method development and iterative improvement in digital settings. AI4AI-Bench and RSIBench-Data assess changes to training algorithms and data strategies through subsequent model performance~\citep{chi2026ai4ai,meng2026rsibenchdata}. RSI-Exam and HarnessDev evaluate executable method or harness revisions~\citep{rsi-exam-2026,wu2026harnessdev}, while PAST-Bench examines the contribution of accumulated experience to later agent performance~\citep{xue2026pastbench}. However, a benchmark jointly evaluating the Physical RSI methods remains an open need.

\begin{table*}[!t]
    \centering
    \caption{Embodied and RSI-related benchmark resources grouped by primary evaluation focus.}
    \label{tab:rsi-benchmark-levels}
    \footnotesize
    \setlength{\tabcolsep}{5pt}
    \renewcommand{\arraystretch}{1.12}
    \begin{tabular}{@{}
        >{\raggedright\arraybackslash}p{0.28\textwidth}
        >{\raggedright\arraybackslash}p{\dimexpr0.72\textwidth-2\tabcolsep\relax}
        @{}}
        \toprule
        \textbf{Evaluation focus} &
        \textbf{Representative resources} \\
        \midrule

        \multicolumn{2}{@{}l@{}}{\textbf{Embodied benchmarks}} \\
        \midrule

        Task Learning, Performance \& Generalization &
        \textbf{Meta-World}~\citep{yu2019metaworld},
        \textbf{RLBench}~\citep{james2020rlbench},
        \textbf{robosuite}~\citep{robosuite2020},
        \textbf{robomimic}~\citep{robomimic2021},
        \textbf{CALVIN}~\citep{mees2022calvin},
        \textbf{LIBERO}~\citep{liu2023libero},
        \textbf{VIMA-Bench}~\citep{pmlr-v202-jiang23b},
        \textbf{BEHAVIOR-1K}~\citep{li2024behavior},
        \textbf{ManiSkill3}~\citep{taomaniskill3},
        \textbf{VLABench}~\citep{zhang2025vlabench},
        \textbf{RoboTwin}~\citep{mu2025robotwin},
        \textbf{EmbodiedBench}~\citep{pmlr-v267-yang25f},
        
        \textbf{RoboCasa365}~\citep{robocasa365}. \\
        \addlinespace[4pt]

        Robustness \& Physical-World Evaluation &
        \textbf{SIMPLER}~\citep{prsi_simpler},
        \textbf{THE COLOSSEUM}~\citep{pumacay2024colosseum},
        
        \textbf{LIBERO-PRO}~\citep{zhou2025liberopro},
        \textbf{LIBERO-Plus}~\citep{fei2025liberoplus},
        \textbf{REALM}~\citep{sedlacek2025realm},
        \textbf{RoboArena}~\citep{pmlr-v305-atreya25a},
        \textbf{RoboDojo}~\citep{chen2026robodojo},
        \textbf{VLA-Arena}~\citep{pmlr-v306-zhang26cp},
        \textbf{RoboTwin 2.0}~\citep{pmlr-v306-chen26j}. \\
        \addlinespace[4pt]

        Feedback \& Evaluation Reliability &
        \textbf{WorldEval}~\citep{li2025worldevalworldmodelrealworld},
        \textbf{FailBench}~\citep{navasardyan2026failbench},
        \textbf{RoboRewardBench}~\citep{lee2026roboreward},
        \textbf{dWorldEval}~\citep{li2026dworldeval},
        \textbf{SC3-Eval}~\citep{tseng2026sc3eval},
        \textbf{RBM-EVAL}~\citep{liang2026robometer}. \\
        \addlinespace[4pt]

        Robot Engineering \& Development &
        \textbf{CaP-Bench (CaP-X)}~\citep{pmlr-v306-fu26u},
        \textbf{RoboCoach-Bench}~\citep{lei2026embocoachbench},
        \textbf{RLE-Bench}~\citep{ma2026rle}. \\

        \midrule

        \multicolumn{2}{@{}l@{}}{\textbf{RSI-related benchmarks (primarily digital)}} \\
        \midrule

         &
        \textbf{MLAgentBench}~\citep{pmlr-v235-huang24y},
        \textbf{MLGym-Bench}~\citep{nathani2025mlgymnewframeworkbenchmark},
        \textbf{MLRC-Bench}~\citep{zhang2025mlrcbench},
        \textbf{RE-Bench}~\citep{pmlr-v267-wijk25a},
        \textbf{AI4AI-Bench}~\citep{chi2026ai4ai},
        \textbf{RSIBench-Data}~\citep{meng2026rsibenchdata}, \textbf{RSI-Exam}~\citep{rsi-exam-2026},
        \textbf{RSIGym}~\citep{meng2026rsigym},
        
        \textbf{OpenRSI Index}~\citep{openrsifoundation2026index}.\\
        \bottomrule
    \end{tabular}
\end{table*}

\section{Challenges and Future Directions}
\label{sec:challenges}
\textbf{Learning from uncertain physical experience.}
Physical interaction provides data and learning signals for robot training, but partial observability, sensor noise, and stochastic outcomes make it difficult to distinguish informative feedback from misleading signals. In Physical RSI, when the system uses uncertain feedback to revise its own training procedure, it may carry misguided updates into later rounds and repeatedly reinforce them through subsequent learning. How to derive reliable learning signals from physical experience so that inherited procedure changes improve subsequent learning without compounding feedback errors remains an open challenge.

\paragraph{Causal attribution across harness components.}
Execution feedback does not uniquely identify the harness components responsible for failure. Dependencies among memory, skills, and workflows allow errors to propagate, so observed deviations may reflect upstream defects or interactions among otherwise valid components. Concurrent updates further confound attribution, as performance changes cannot be assigned to individual modifications without controlled comparisons. Partial observability and limited physical trial budgets restrict such comparisons, leaving both the source of failure and the effect of intervention uncertain. Effective harness adaptation therefore depends on distinguishing causal dependencies from observed associations and isolating the effects of modifications sufficiently to justify their targets and scope.

\paragraph{Non-stationarity in the improvement loop.}
Evidence supporting a retained update is conditional on the system configuration and physical conditions under which it was obtained. Subsequent component revisions can alter dependencies and invalidate compatibility assumptions, while environmental changes can shift the conditions governing an update's effectiveness. Historical outcomes consequently provide uneven support for current adoption decisions. The same dependence applies to improvement strategies, whose measured effectiveness may change with the capabilities and experience available to the system. Sustained improvement requires tracking these dependencies and reassessing evidence as its underlying conditions change, while limiting the physical cost of revalidation.

\paragraph{Reward hacking and evaluation leakage.}
Improvement procedures in Physical RSI rely on success detectors, learned rewards, and model-based judgments that may not faithfully reflect physical task outcomes. Benchmarks of robotic outcome assessment expose errors in these signals \citep{navasardyan2026failbench,prsi_nofree}, while reward hacking in digital research agents illustrates how optimization can exploit an imperfect evaluation target \citep{huang2026rewardhacking}. As $\Phi$ evolves, changes to $\mathrm{Train}$ or $\mathrm{Ver}$ may amplify such biases across subsequent updates. Moreover, even when the external evaluator $E$ and criterion $Q$ are protected from direct access, repeated use of evaluation results for candidate selection can compromise their statistical independence \citep{dwork2015generalization}. A central direction is therefore to combine independently checked physical outcomes with controlled disclosure of evaluation feedback and fresh or statistically corrected final tests, so that higher scores reflect actual capability gains rather than exploitation of the measurement process.

\paragraph{Safety across recursive updates.}
Maintaining safety becomes harder as $\Phi$ evolves, particularly when drift in $\mathrm{Ver}$ favors updates that pass local checks while increasing physical risk. Existing guarantees remain limited: the proof-preserving methods reviewed in Section~\ref{sec:safety_control} cover structured model and controller revisions~\citep{fulton2019offmodel}. Extending these guarantees to inherited changes in $\Phi$ remains open. Exhaustive revalidation after every update can exceed physical trial budgets and itself incur harm. A concrete direction is selective reassessment: identifying assumptions invalidated by an update or environmental drift, reusing evidence that remains valid, and concentrating new trials on unresolved safety claims. Such reassessment requires per-round risk bounds to remain valid as verification and data collection adapt, while preserving protected requirements and effective interruption. Comparisons of successive procedures from matched starting systems and experience, under equal budgets, fixed safety rules, and protected $E,Q$, would test whether recursive gain can be established across tasks without relaxing constraints or concealing physical risk incurred during rejected trials, resets, and recovery.

\paragraph{Benchmarking verified recursive gain.}
Existing robot benchmarks assess task performance, feedback reliability, and agent-driven program or training development \citep{navasardyan2026failbench,pmlr-v306-fu26u,lei2026embocoachbench}, but do not directly establish whether revisions to the improvement procedure $\Phi$ make subsequent improvement more effective. Higher successor performance may result from accumulated experience or stronger starting systems rather than a better improvement mechanism. The central challenge is to evaluate mechanism-level recursive gain under comparable conditions. Future benchmarks should track autonomous revisions, inheritance, and subsequent use of $\Phi$, and compare $\Phi_k$ with $\Phi_{k-1}$ from the same starting generation and experience, under matched tasks and resource budgets.

\section{Conclusion}\label{sec:conclusion}

This study introduces a unified framework for Physical Recursive Self-Improvement, extending robot autonomy from task execution to improving both capabilities and the procedures that produce them. The L0--L5 framework classifies systems according to their update targets, task scope, and evidence of recursive gain. Guided by this framework, we investigate existing methods around harness adaptation and model training, alongside the safety, control, and evaluation mechanisms that support successive updates. Our analysis highlights a key gap: improved task performance alone does not establish that the improvement procedure itself has become more effective. Demonstrating Physical RSI therefore requires independent comparisons of successive procedures under matched starting conditions and resource budgets, with gains demonstrated across tasks. Future progress depends on reliable physical feedback, clearer attribution of gains to procedure changes, and safety assessments that remain valid as systems and environments evolve. Addressing these challenges would pave the way for robots that autonomously improve both what they can do and how they learn, while ensuring safety. The ultimate goal of Physical RSI is for robots to become increasingly capable of improving themselves as they encounter new tasks and challenges in the physical world.

\bibliographystyle{plainnat}
\bibliography{ref_hai,verification,agent/references, training/training,intro,Safety_and_Control/safety_control_all_references,Evaluation_of_Physical_RSI/evaluation_physical_rsi_references,foundation/foundation}

\end{document}